\documentclass[11pt,a4paper]{article}

\usepackage[T1]{fontenc}
\usepackage[utf8]{inputenc}
\usepackage{lmodern}
\usepackage{amsmath,amssymb,amsthm}
\usepackage{graphicx}
\usepackage[numbers]{natbib}
\usepackage{hyperref}
\usepackage{xcolor}
\usepackage{booktabs}
\usepackage{enumitem}
\usepackage{microtype}
\usepackage[margin=1in]{geometry}
\usepackage{float}
\usepackage[labelformat=simple]{subfig}

\graphicspath{{figures/}}

\title{Structured Fluctuations and the Information Dynamics of Self-Maintenance in Growing Neural Cellular Automata}

\author{
  Atsushi Masumori$^{1,2,3,*}$ \quad Hiroki Sato$^{1,2,3}$ \quad Takashi Ikegami$^{1,2,3}$ \\[6pt]
  {\small $^{1}$\,The University of Tokyo \quad $^{2}$\,Atomi University \quad $^{3}$\,Alternative Machine Inc.} \\
  {\small $^{*}$\,Correspondence: \texttt{atsushi.masumori@gmail.com}}
}

\date{}
\begin{document}
\maketitle

\begin{abstract}
Growing Neural Cellular Automata (GNCA) are capable of robust self-maintenance and self-repair, yet the internal dynamical mechanisms that support these capabilities remain poorly understood. Here, we investigate the role of internal fluctuations---temporal micro-variability of hidden channel states---in a trained GNCA model, challenging the assumption that such variability is merely residual stochastic noise. Through systematic analysis spanning update-rate sweeps, spatial correlation measurements, dimensionality reduction of collective state trajectories, localized damage experiments, transfer entropy vector field estimation, and partial information decomposition, we show that internal fluctuations are spatially structured, dynamically coupled to an attracting collective state, and associated with distributed small-magnitude updates that contribute to damage recovery. Damage induces a global deviation in latent state space followed by gradual re-convergence, and suppressing distributed small-magnitude updates associated with baseline fluctuation dynamics outside a permissive radius that encompasses the majority of the cells significantly impairs recovery. Transfer entropy analysis characterizes a spatially differentiated repair response: corrective inward flow near the damage site coexists with outward perturbation propagation at greater distances. Partial information decomposition further suggests a regime shift from synergy-dominant resting computation to redundancy-increased coordination during recovery. These findings indicate that GNCA self-repair emerges from high-dimensional nonlinear collective dynamics in which internal fluctuations serve as a functional component supporting information flow, coordination, and return toward an attracting recurrent state.
\end{abstract}


\bigskip

\section{Introduction}
\label{sec:introduction}

Biological organisms exhibit a remarkable capacity for self-maintenance and self-repair, sustaining coherent form and function through the continuous coordination of decentralized cellular processes.
Understanding how such global robustness arises from local interactions remains a central challenge across developmental biology, complex systems science, and artificial life \citep{turing1952chemical,kauffman1993origins,levin2021bioelectric}.
Growing Neural Cellular Automata (GNCA) \citep{mordvintsev2020growing} provide a tractable computational framework for studying these phenomena.
Since their introduction, NCA have been extended to isotropic architectures \citep{mordvintsev2022isotropic}, texture synthesis \citep{niklasson2021selforg,mordvintsev2021texture,pajouheshgar2023dynca}, self-classification \citep{randazzo2020selfclass}, intrinsic motivation \citep{grasso2022empowered}, variational generative modeling \citep{palm2022variational}, and action generation in embodied settings \citep{variengien2021selforganized,horibe2021regenerating,sato2024morphomotor}.
These extensions demonstrate that a learned local update rule applied uniformly across a spatial grid can give rise to morphogenesis, pattern maintenance, and recovery from perturbation---all without centralized control.
Although GNCA are not biologically realistic models of tissue repair, they provide a tractable artificial system that shares several organizational constraints with developmental and regenerative systems, including local interaction, decentralized control, morphology maintenance, and recovery from perturbation. For the present study, this makes GNCA useful as an experimentally accessible system in which these features can be analyzed together.

While previous NCA studies have primarily focused on emergent patterns, the internal dynamical structure that maintains these patterns has remained largely unexplored.
Recent work has begun to investigate the relationship between NCA architecture and emergent dynamics \citep{xu2024emergent,catrina2024learning}, but these studies focus on the dynamical properties of the output patterns rather than the internal state dynamics that sustain them.
The multi-channel hidden states have typically been treated as an opaque computational substrate.

Yet a distinctive feature of GNCA dynamics is that each cell maintains a high-dimensional internal state that evolves continuously over time, and even after a stable morphological pattern has been achieved, these internal states are not static; they exhibit persistent temporal variability that we term \textit{fluctuations}.
This micro-variability might initially be dismissed as computational noise or an artefact of the stochastic update schedule.
However, in biological systems, fluctuations at the molecular and cellular level are increasingly recognized not as mere noise but as functional components that support information processing, cellular decision-making, and adaptive coordination \citep{elowitz2002stochastic,eldar2010functional,huang2009heterogeneity}.
In gene regulatory networks, cell-to-cell variability reflects the multi-stable attractor structure of the underlying dynamics rather than random perturbation \citep{huang2005cellfates}.
At a more abstract level, Kaneko and Ikegami's concept of \textit{homeochaos}---dynamic stability sustained through weak high-dimensional chaos---suggests that structured variability can itself be the mechanism by which complex systems maintain robustness, rather than a threat to it \citep{kaneko1992homeochaos}.
These diverse precedents motivate the hypothesis that internal fluctuations in GNCA may play a similarly functional role.

Testing this hypothesis requires two complementary perspectives: a dynamical systems perspective to determine whether the collective behavior of the system is organized around low-dimensional structures such as attractors or invariant manifolds, and an information-theoretic perspective to quantify how information is routed, shared, and transformed among cells during maintenance and repair.
The dynamical systems approach is well established for spatially extended nonlinear systems, where dimensionality reduction and correlation analysis have long been used to identify collective modes and characterize the geometry of high-dimensional state spaces.

On the information-theoretic side, an extensive body of work on classical cellular automata (CA) provides a mature toolkit.
Lizier, Prokopenko, and colleagues decomposed distributed computation into information storage, transfer, and modification, applying transfer entropy \citep{schreiber2000measuring} to identify directed information flow and coherent computational structures in CA \citep{lizier2008local,lizier2012local,lizier2014framework}.
Partial information decomposition (PID) \citep{williams2010nonnegative} and its temporal extension $\Phi$ID \citep{mediano2025integrated} further enable decomposition of multi-source and multi-channel information into redundant, unique, and synergistic components.
These methods have been applied as spatiotemporal filters in CA \citep{flecker2011pid,finn2018pointwise} and to characterize regime shifts in neural systems \citep{luppi2024information}.
To our knowledge, few studies have applied these approaches to the internal dynamics of trained Neural Cellular Automata, leaving fundamental questions unanswered: what is the collective dynamical regime of the system, how does information flow between cells during self-repair, and are multi-channel hidden states organized as independent features, redundant copies, or a synergistic computational substrate?

In this work, we present a comprehensive empirical investigation of internal fluctuation dynamics in a trained GNCA model, bringing together dynamical systems analysis and information-theoretic decomposition to address these questions.
Note that our focus is on the \textit{post-developmental} regime: rather than analyzing the morphogenetic growth process itself, we examine the dynamics of the system after it has reached its target morphology, during both steady-state self-maintenance and recovery from localized damage.
Our central hypothesis is that self-maintenance and self-repair in GNCA emerge from high-dimensional nonlinear collective dynamics, where internal fluctuations are not merely noise but a functional component supporting information flow, spatial coordination, and recovery.
This perspective draws on the theoretical traditions of coupled map lattice dynamics \citep{kaneko1992overview,kaneko2001complex}, attractor-based models of biological robustness \citep{waddington1957strategy,huang2005cellfates}, and homeostatic regulation in adaptive systems \citep{ashby1960design,dipaolo2005autopoiesis}.
We organize our analysis around three interrelated questions.
First, what is the structure of internal fluctuations, and how do they relate to the collective dynamical state of the system?
Second, how does the system respond to localized damage at both local and global scales, and what role do fluctuations play in recovery?
Third, what information-theoretic signatures characterize the computational regime of the system during resting maintenance versus active repair?

Our results progressively demonstrate that (i) fluctuations are spatially structured collective phenomena intrinsic to the learned dynamics, (ii) the system exhibits attracting recurrent organization, with damage inducing displacement from and return toward this recurrent state, (iii) distributed small-magnitude updates associated with baseline fluctuation dynamics contribute to recovery, (iv) repair triggers a spatially differentiated information flow pattern combining local correction with global broadcasting, and (v) the computational regime shifts from synergy-dominant (integrative, nonlinear combination of inputs) to redundancy-enhanced (overlapping, error-tolerant signaling) during recovery.

\section{Methods}
\label{sec:methods}

\subsection{Growing Neural Cellular Automata}
\label{sec:model}

GNCA \citep{mordvintsev2020growing} are continuous-state cellular automata in which every cell on a two-dimensional grid updates its state according to a shared, learned local rule parameterized by a small neural network.
Starting from a single seed cell, the system self-organizes to reproduce a target image through spatio-temporal development, and---when trained with periodic damage---can regenerate the target pattern after localized destruction.
We briefly describe the architecture and training procedure below; full details can be found in \citet{mordvintsev2020growing}.

\vspace{6pt}\noindent\textbf{State representation.}
Each cell carries a state vector of $C = 16$ continuous channels.
The first four channels encode RGBA (red, green, blue, alpha) values that are compared against the target image during training; the remaining 12 channels are \textit{hidden state} channels with no predefined semantics, which the model is free to use as an internal communication and memory substrate.
The alpha channel (channel index 3) additionally serves as a cell viability indicator.
The full system state at time $t$ is a tensor $\mathbf{S}(t) \in \mathbb{R}^{C \times H \times W}$ with grid dimensions $H = W = 72$ in our experiments.

\vspace{6pt}\noindent\textbf{Perception.}
At each time step, each cell perceives its $3 \times 3$ local neighborhood through three fixed convolutional filters applied independently to each of the 16 state channels: an identity filter (extracting the cell's own value), a horizontal Sobel filter (estimating $\partial s / \partial x$), and a vertical Sobel filter (estimating $\partial s / \partial y$).
This produces a 48-dimensional perception vector ($16 \times 3$ filters) per cell.

\vspace{6pt}\noindent\textbf{Update rule.}
The perception vector is processed by a shared two-layer neural network implemented as $1 \times 1$ convolutions: a first layer mapping from 48 to 128 units with ReLU activation, followed by a second layer mapping from 128 to 16 units (approximately 8{,}000 learnable parameters in total).
The weights of the final layer are initialized to zero, ensuring that the model begins with an identity (no-update) behavior.
The network output is treated as a \textit{residual update} $\Delta \mathbf{s}$ added to the current state: $\mathbf{s}(t{+}1) = \mathbf{s}(t) + \Delta \mathbf{s}(t)$.

\vspace{6pt}\noindent\textbf{Alive masking.}
Before and after each update, an alive mask is applied: a cell is considered alive if any cell in its $3 \times 3$ neighborhood has an alpha channel value exceeding 0.1 (implemented via max-pooling with circular boundary conditions).
Dead cells have their entire state vector set to zero, preventing spurious growth in empty regions.

\vspace{6pt}\noindent\textbf{Stochastic cell update.}
Not all cells are updated at every time step.
An update rate parameter $\text{UR} \in (0, 1]$ controls the probability that each cell applies its computed update; with probability $1 - \text{UR}$, the cell retains its previous state unchanged across all channels.
This mechanism breaks the assumption of a global synchronous clock and forces the learned dynamics to be robust to asynchronous updating.
In the original GNCA, $\text{UR} = 0.5$; in this study, we systematically vary UR from 0.1 to 1.0 to examine its effect on internal fluctuation dynamics, and use $\text{UR} = 0.8$ as the default for most analyses.
We note that qualitatively similar results were obtained with an alternative implementation in which the stochastic mask was applied independently per channel rather than per cell, indicating that the phenomena reported below are robust to the granularity of the update mechanism.

\begin{figure}[!htb]
  \centering
  \includegraphics[width=\linewidth]{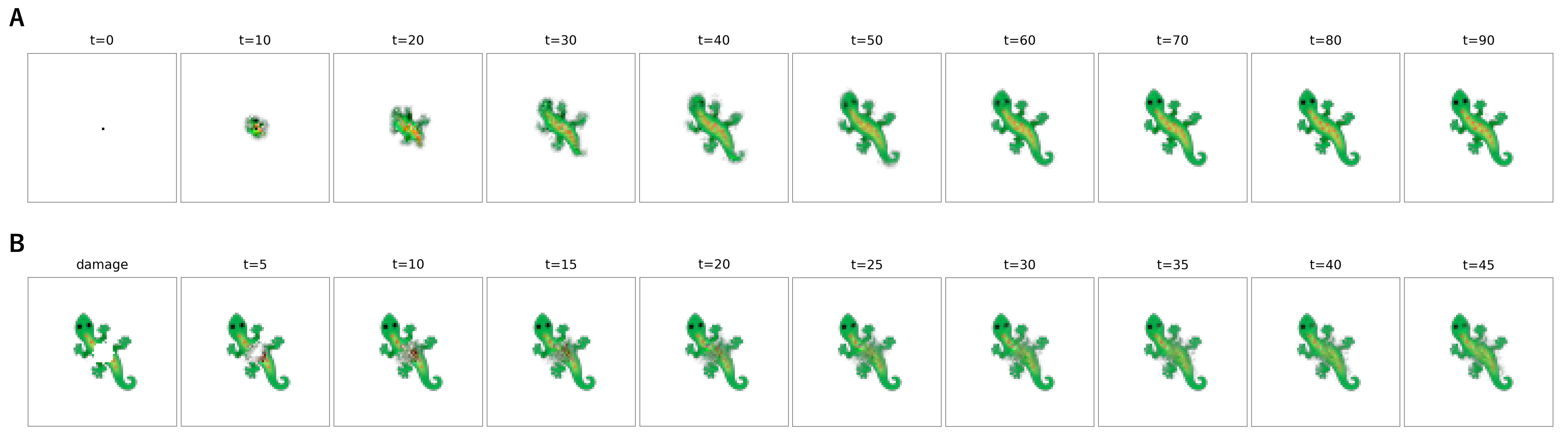}
  \caption{(\textbf{A}) Developmental process of the GNCA lizard morphology from a single seed cell, shown at 10-step intervals ($t = 0$--$90$). The system self-organizes to reproduce the target pattern within approximately 50 steps.
  (\textbf{B}) Regeneration after localized damage, shown at 5-step intervals ($t = 5$--$45$). The damaged region is progressively restored through coordinated cell dynamics.
  \label{fig:develop_regeneration}}
\end{figure}

\vspace{6pt}\noindent\textbf{Target image and training.}
The target image is the lizard emoji from the Google Noto Emoji set (Figure~\ref{fig:develop_regeneration}), resized to $40 \times 40$ pixels and centered within the $72 \times 72$ grid with zero-padding.
Training follows the ``sample pool'' algorithm and the regenerative training regime of \citet{mordvintsev2020growing}: a pool of partially grown states is maintained, and at each training iteration a batch of 8 samples is drawn from the pool, advanced for a random number of steps (64--96), and the MSE loss between the RGBA channels and the target is backpropagated through the unrolled computation graph.
To encourage regeneration, a fixed number of samples in each batch are damaged by applying random circular masks before the forward pass, and the sample with the highest loss is replaced with the initial seed state after each iteration, forcing the model to learn both growth from a single seed cell and recovery from partial destruction.
Circular boundary conditions (toroidal grid) are used for both the perception and alive-masking operations.
Models were trained for 10,000 epochs using Adam with learning rate $2\times10^{-3}$ and a persistent sample pool of 1024 states. After training, approximately 500 cells were alive in the steady-state morphology (Figure~\ref{fig:develop_regeneration}A).
Figure~\ref{fig:develop_regeneration}B illustrates the regeneration process after localized damage.
We analyzed eight independently trained models initialized with different random seeds; the random seed for each run was recorded with the corresponding checkpoint.

\subsection{Analysis Overview}
\label{sec:experimental_design}

All analyses focus on the post-developmental regime: models are first run for a relaxation period (typically 1000 steps) until the target morphology is stably maintained, and subsequent dynamics are recorded.
We organize our experiments into three stages.

The first stage characterizes the basic properties of internal fluctuations (\S\ref{sec:results_fluctuations}--\S\ref{sec:results_cores}).
We compute per-cell temporal statistics (mean, standard deviation, activity) across update rates, analyze spatiotemporal correlations via spatial autocorrelation and temporal autocorrelation functions, and map the spatial distribution of fluctuation intensity across channels.

The second stage investigates collective dynamics and the functional role of fluctuations (\S\ref{sec:results_attractor}--\S\ref{sec:results_damage}).
System-level state trajectories are visualized via UMAP \citep{mcinnes2018umap} to characterize their recurrent structure.
Localized damage is applied by resetting all 16 channel values to zero within a circular region centered at a fixed coordinate within the alive cell area; the specific parameters (radius, frequency, number of cycles) vary across experiments and are described in the corresponding results sections.
Recovery is quantified using two complementary metrics: the cosine distance from a pre-damage baseline state (see Appendix~\ref{app:metrics} for formal definition) and the mean squared error (MSE) of RGB channels within the damage region.
A fluctuation suppression experiment tests their functional contribution by selectively blocking micro-fluctuations outside a permissive radius while allowing larger state changes to propagate freely.

The third stage applies information-theoretic methods to characterize the computational regime (\S\ref{sec:results_information}).
Transfer entropy (TE) \citep{schreiber2000measuring} is used to construct spatial vector fields of directed information flow under undamaged and post-damage conditions.
Inter-channel coupling is analyzed via integrated information decomposition ($\Phi$ID) \citep{mediano2025integrated}, decomposing temporal mutual information between channel pairs into redundancy, unique information, and synergy.
Spatial computation is characterized via partial information decomposition (PID) \citep{williams2010nonnegative}, using spatial gradient and center state features as sources and the local state update as target.
All information-theoretic quantities use 8-bin quantile discretization; formal definitions are provided in Appendix~\ref{app:metrics}.
These quantities are used as descriptive characterizations of directed predictive structure and information organization, rather than as evidence of direct causal transmission.

\section{Characterization of Structured Fluctuations}
\label{sec:results_fluctuations}

A prerequisite for understanding the role of fluctuations in GNCA is to establish their basic properties: how they depend on system parameters, whether they represent transient or persistent phenomena, how they vary across the cell population, and whether they exhibit spatial organization.
We address each of these questions in turn.

\subsection{Update Rate Dependency and Dynamical Steady State}
\label{sec:results_stats}

For each alive cell at position $(i,j)$, we computed the temporal mean $\bar{s}_{c,i,j} = \frac{1}{T}\sum_{t=1}^{T} s_{c,i,j}(t)$ and temporal standard deviation $\sigma_{c,i,j} = \sqrt{\frac{1}{T}\sum_{t=1}^{T}(s_{c,i,j}(t) - \bar{s}_{c,i,j})^2}$ over $T$ steady-state time steps for each channel $c$.

\begin{figure}[!htb]
  \centering
  \includegraphics[width=\linewidth]{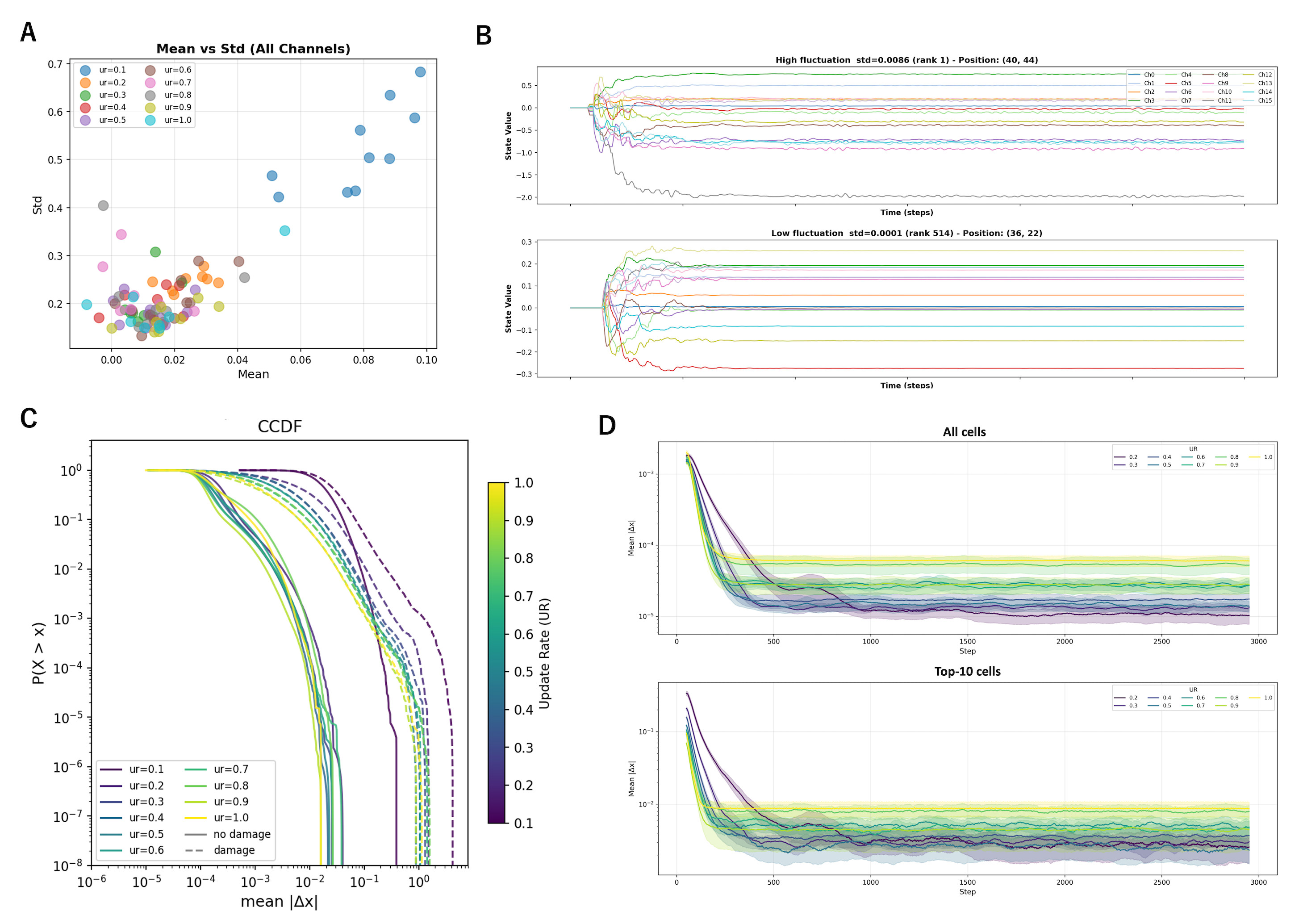}
  \caption{(\textbf{A}) Scatter plot of mean internal state magnitude versus temporal fluctuation (standard deviation) for update rates UR $\in [0.1, 1.0]$. Each point represents one model (averaged across all 16 channels) at a given update rate. Excluding UR~$= 0.1$ (where training was unstable), data points cluster together regardless of update rate, indicating that fluctuations are an intrinsic property of the learned dynamics rather than a consequence of stochastic updating.
  (\textbf{B}) Raw internal state time series (all 16 channels) for representative cells at UR $= 0.8$. Low-fluctuation cells converge rapidly to stable values, while high-fluctuation cells exhibit persistent, structured oscillations across channels, indicative of ongoing dynamical activity rather than noise.
  (\textbf{C}) Complementary cumulative distribution functions (CCDFs) of per-cell activity under no-damage (solid lines) and post-damage (dashed lines) conditions for update rates UR $= 0.1$--$1.0$ (color scale). The log-normal distribution provides a significantly better fit than a power law across all conditions ($\Delta$AIC $\ll 0$).
  (\textbf{D}) Time series of activity $\|\Delta\mathbf{s}\|$ at UR $= 0.8$ (smoothed with a 20-step sliding window). Top: mean across all alive cells ($\sim 10^{-4}$). Bottom: top-10 highest-activity cells ($\sim 10^{-2}$), revealing two orders of magnitude difference in fluctuation intensity across the cell population.
  \label{fig:basic_characters}}
\end{figure}

Averaging these quantities across all 16 channels for each model ($n = 8$) and plotting across update rates ranging from UR~$= 0.1$ to UR~$= 1.0$ revealed a striking pattern (Figure~\ref{fig:basic_characters}A).
Excluding UR~$= 0.1$, where training was unstable, the data points clustered together regardless of update rate, with comparable mean state magnitudes and temporal standard deviations across the range UR~$= 0.2$--$1.0$.
This clustering indicates that fluctuations are an intrinsic property of the learned dynamics rather than a consequence of the stochastic update schedule: even at UR~$= 1.0$, where every cell is updated at every step and stochastic update-induced variability is absent, substantial fluctuations persist.

Visualization of raw internal state time series for individual cells confirmed that fluctuations persist well after the system reaches its target morphology (Figure~\ref{fig:basic_characters}B).
Low-fluctuation cells converged to near-constant values across all 16 channels (temporal standard deviation $\sim 10^{-4}$), while high-fluctuation cells exhibited persistent, structured oscillations approximately two orders of magnitude larger ($\sim 10^{-2}$), indicating substantial heterogeneity in dynamical activity across the cell population.

The statistical structure of activity---defined as the step-to-step state change magnitude---provides further insight into the nature of the fluctuations.
The complementary cumulative distribution function (CCDF) of per-cell activity in the steady state was well characterized by a log-normal distribution across all update rates (Figure~\ref{fig:basic_characters}C).
Model comparison via the Akaike Information Criterion strongly favored the log-normal over a power-law fit ($\Delta\text{AIC} \ll 0$), indicating that activity fluctuations arise from multiplicative rather than scale-free processes.
The mode of the distribution shifted systematically toward higher values with increasing update rate, consistent with the update-rate dependence observed in the preceding analyses.

Following damage, the CCDF curves shifted rightward relative to the undamaged baseline (Figure~\ref{fig:basic_characters}C, dashed versus solid lines), indicating an increase in large-magnitude fluctuations.
This damage-induced broadening was present across all update rates but was most pronounced at low update rates (UR~$\leq 0.3$), where the undamaged distribution was narrow and the damage-induced tail extended over several orders of magnitude.
At higher update rates, both distributions were broader and the relative shift was smaller, suggesting that models with more frequent updates are inherently more variable and thus less sensitive to perturbation in relative terms.

The time series of step-to-step state change magnitude further revealed the heterogeneous nature of the fluctuations (Figure~\ref{fig:basic_characters}D).
The population-averaged activity was on the order of $10^{-4}$, but the top-10 most active cells exhibited magnitudes on the order of $10^{-2}$---two orders of magnitude larger---indicating that fluctuation intensity is concentrated in a small subset of cells rather than uniformly distributed.

In summary, GNCA exhibit persistent micro-variability that is intrinsic to the learned dynamics rather than an artefact of stochastic updating, and damage amplifies this variability across the cell population.

\subsection{Spatiotemporal Correlations}
\label{sec:results_correlation}

To quantify the spatial extent of dynamical coupling, we computed the radially averaged spatial autocorrelation function $C(r)$ of the internal state field and extracted a characteristic correlation length $\xi$ by fitting an exponential decay model $C(r) \propto \exp(-r/\xi)$.

\begin{figure}[!htb]
  \centering
  \includegraphics[width=\linewidth]{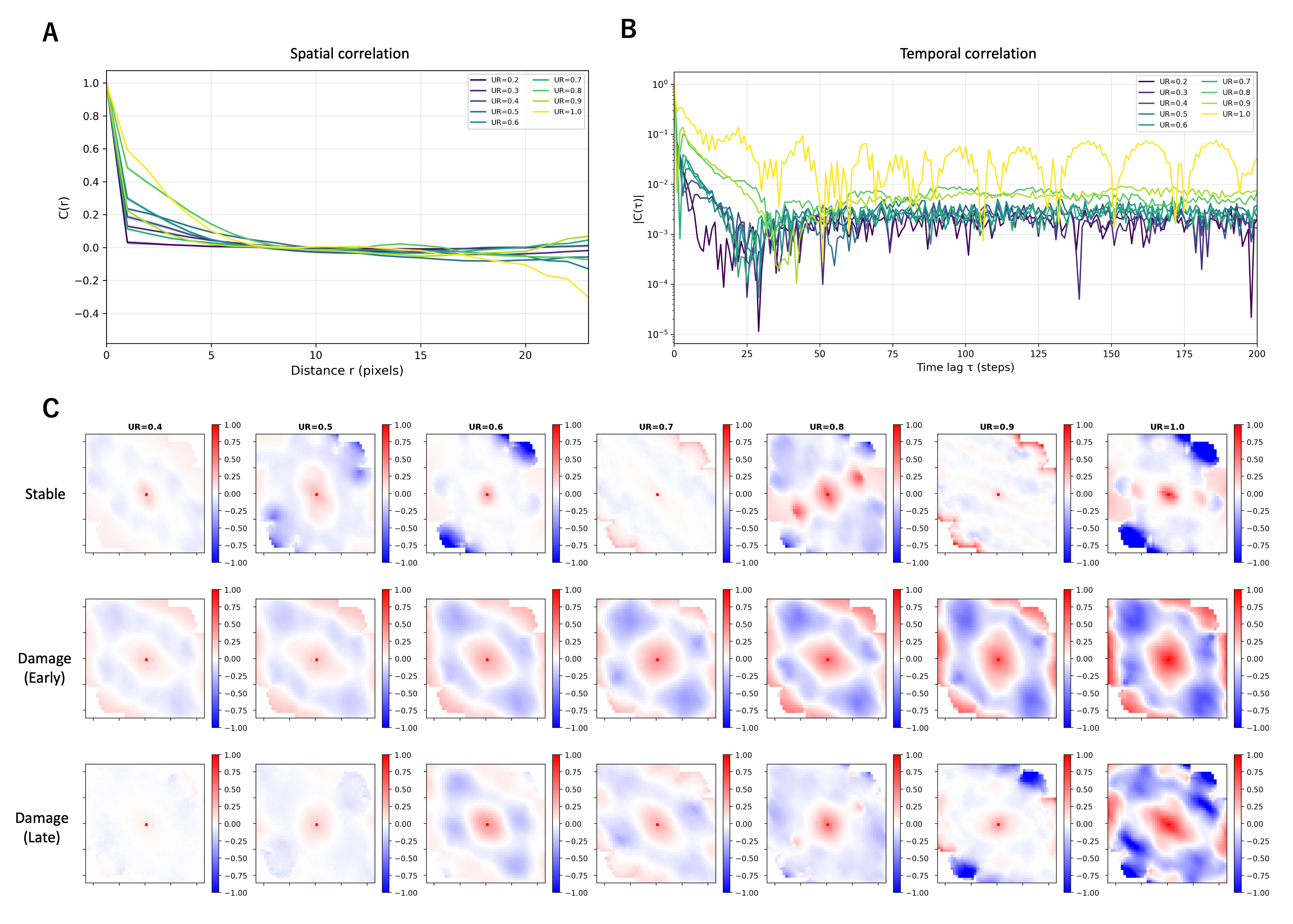}
  \caption{Spatiotemporal correlation structure of internal fluctuations. (\textbf{A}) Radially averaged spatial autocorrelation function $C(r)$. The correlation length increases with UR, indicating that more frequent updates support longer-range spatial coupling. (\textbf{B}) Temporal autocorrelation function ACF($\tau$) of per-cell activity $\|\Delta\mathbf{s}\|$. Higher update rates show stronger temporal persistence, with all conditions settling to non-zero levels by $\tau \approx 25$. At UR~$= 1.0$, oscillatory autocorrelation suggests a strong periodic component in the absence of stochastic updating. (\textbf{C}) Two-dimensional spatial autocorrelation maps in displacement space $(\Delta x, \Delta y)$, revealing the anisotropy and spatial extent of dynamical coupling across update rates.\label{fig:correlation}}
\end{figure}

The resulting profiles revealed that internal state dynamics are not spatially independent but exhibit extended correlations (Figure~\ref{fig:correlation}A).
Across all update rates, $C(r)$ decayed approximately exponentially with distance, reaching zero by $r \approx 20$ pixels; beyond $r \approx 23$ pixels no alive-cell pairs remained for estimation.
The correlation length $\xi$ increased systematically with update rate: at low update rates (UR~$= 0.2$), $\xi$ was relatively short, while at UR~$= 1.0$ the correlation length expanded substantially.
Notably, $\xi$ exceeded the direct perception radius of one pixel (the $3 \times 3$ kernel) across all update rates, indicating that the learned dynamics create emergent spatial coupling beyond nearest-neighbor interactions through multi-step propagation.
The correlation length thus represents the effective coordination scale of the system: the spatial extent over which cells participate in coordinated dynamical fluctuations.
This reflects a property of the learned dynamics rather than a trivial consequence of the local interaction range.
The two-dimensional spatial autocorrelation maps, plotted in displacement space $(\Delta x, \Delta y)$ rather than absolute spatial coordinates (Figure~\ref{fig:correlation}C), confirmed and extended the radial analysis.
Across all update rates, the maps showed elevated positive correlation near the center with positive or negative correlations appearing at the edges of the alive region.
At UR~$= 0.8$ and $1.0$, locally elevated correlations also emerged at intermediate distances, suggesting the formation of mesoscale spatial structure in the fluctuation dynamics beyond simple nearest-neighbor coupling.
Following damage, the 2D spatial ACF reorganized into a consistent pattern across all update rates: positive correlation at the center, negative correlation in the surrounding ring, and positive correlation again at the edges.
This structured pattern contrasts with the UR-dependent variability observed in the resting state, suggesting that damage imposes a common spatial correlation structure regardless of the underlying update rate.
In the late recovery window, the correlation maps shifted back toward the resting configuration, though full recovery was not complete within the observation period.

The temporal autocorrelation of per-cell activity (Figure~\ref{fig:correlation}B) decayed gradually across all update rates, settling to a non-zero fluctuating level by approximately $\tau \approx 25$ steps, indicating persistent long-range temporal correlations.
Higher update rates yielded consistently higher autocorrelation values, reflecting stronger temporal persistence in the fluctuation dynamics.
At UR~$= 1.0$, where stochastic cell updating is absent, the autocorrelation exhibited clear oscillatory behavior, suggesting  a strong periodic component in the deterministic dynamics.
This is consistent with the recurrent trajectory structure revealed by the UMAP analysis in \S\ref{sec:results_umap}.

\section{Emergence of Fluctuation Cores}
\label{sec:results_cores}

\begin{figure}[!htb]
  \centering
  \includegraphics[width=\linewidth]{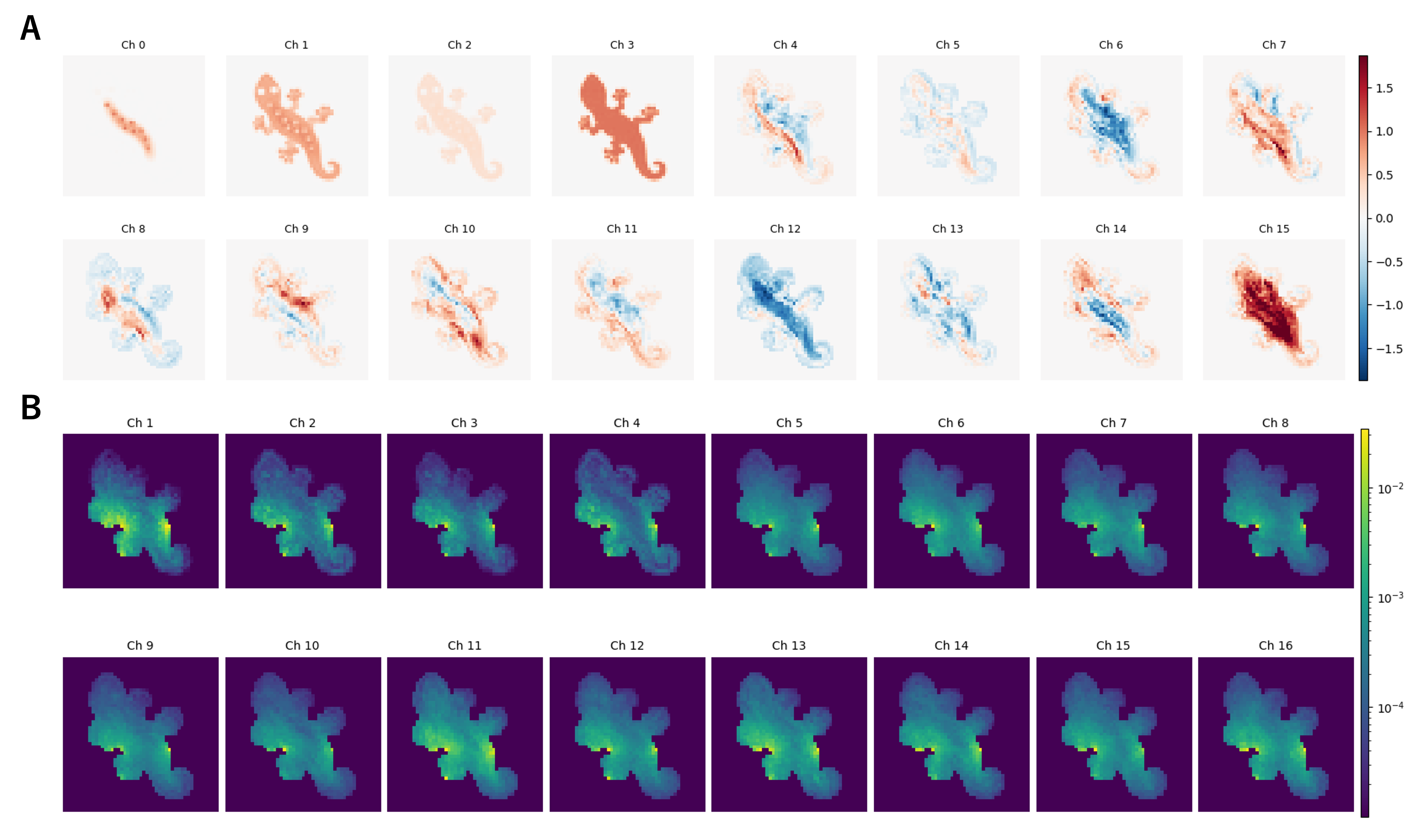}
  \caption{Spatial distributions across channels at UR $= 0.8$. (\textbf{A}) Mean state maps exhibit strong channel-specific spatial patterns reflecting representational specialization. (\textbf{B}) Temporal fluctuation maps show fluctuation cores at similar spatial locations across all 16 channels, suggesting a shared dynamical infrastructure. The contrast between fluctuation and mean-state structure indicates that micro-variability arises from collective dynamical properties rather than channel-specific encoding alone.\label{fig:spatial_maps}}
\end{figure}

Mean internal states and temporal fluctuations showed contrasting spatial organization across channels (Figure~\ref{fig:spatial_maps}).
The mean internal state maps exhibited strong channel-specific spatial patterns, suggesting representational differentiation among the 16 channels (Figure~\ref{fig:spatial_maps}A).
In contrast, temporal fluctuations were concentrated in a small number of localized regions that appeared at similar spatial locations across channels (Figure~\ref{fig:spatial_maps}B).
We refer to these localized regions of elevated micro-variability as ``fluctuation cores.''
In rare cases, the fluctuation distribution was more diffuse, spreading broadly across the grid without forming distinct cores; however, such instances were uncommon.
Notably, the tendency to form a small number of spatially localized fluctuation cores was observed across update rates and training runs, and qualitatively similar fluctuation-core-like structures were also found in additional target morphologies 
(Appendix~\ref{app:analysis}). This suggests that the emergence of localized fluctuation cores is not merely an artefact
of a particular trained instance or target configuration.

Interestingly, fluctuation cores were not distributed uniformly across the organism.
They rarely appeared in the central bulk region and instead tended to emerge near morphological boundaries and protrusions.
This spatial bias suggests that fluctuation cores are not arbitrary sites of internal fluctuation, but are coupled to the geometry of the maintained morphology itself.
One possible interpretation is that boundary and protrusion regions impose stronger local dynamical constraints and larger state sensitivities, making them preferred sites for the amplification of small perturbations and the generation of sustained micro-variability.

Fluctuation cores appeared at similar spatial locations across channels, despite the channel-specific organization of mean internal states.
The channel-invariant nature of fluctuation cores suggests that they reflect a shared dynamical infrastructure rather than channel-specific representational features.
These regions may correspond to sites of heightened dynamical sensitivity, where small state variations are amplified and propagated through the learned local update dynamics.

\begin{figure}[!htb]
  \centering
  \includegraphics[width=\linewidth]{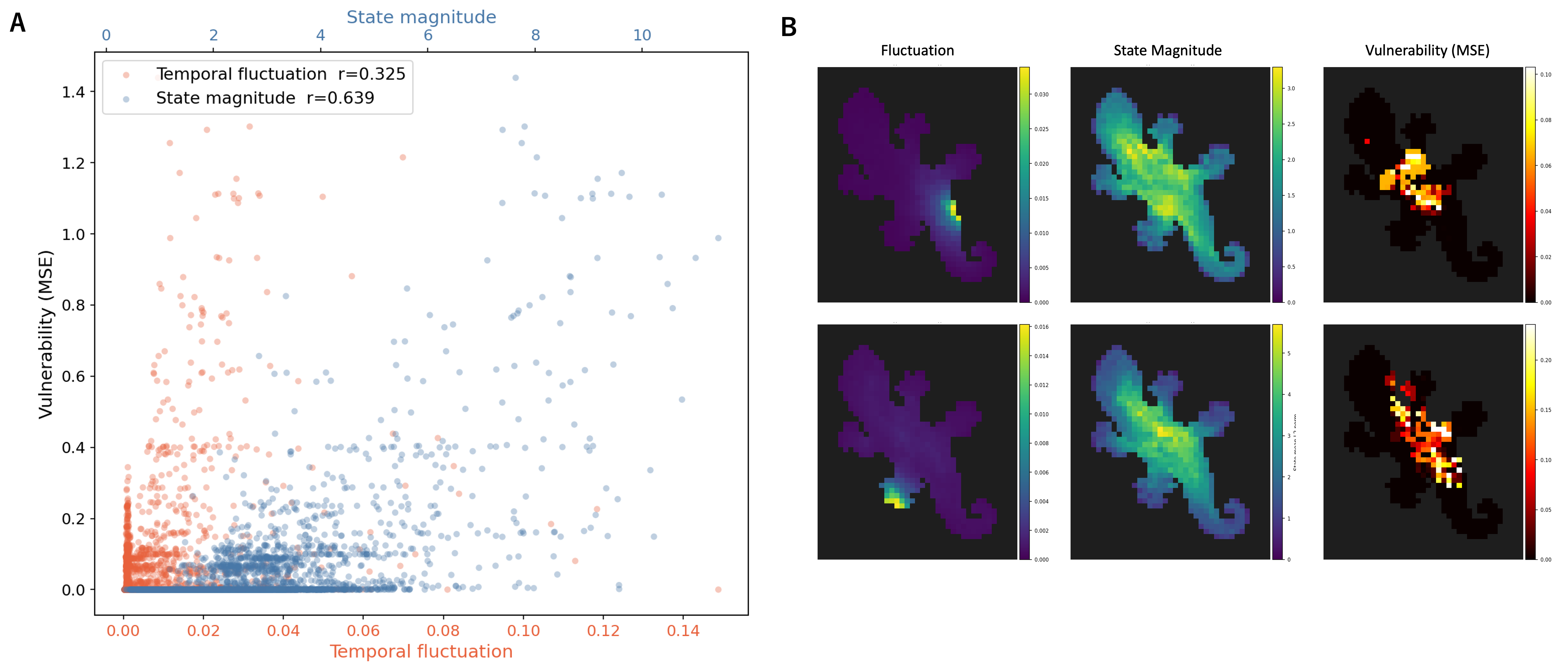}
  \caption{Cell-wise damage vulnerability as a function of temporal fluctuation and state magnitude. (\textbf{A}) Each cell is independently perturbed and the resulting recovery error (MSE) is measured. Vulnerability correlates more strongly with state magnitude ($r = 0.639$) than with temporal fluctuation ($r = 0.325$), indicating that fragility is primarily determined by internal state magnitude. (\textbf{B}) Spatial maps confirm that highly vulnerable regions overlap with regions of large state magnitude, with weaker correspondence to fluctuation cores.\label{fig:vulnerability}}
\end{figure}

Given that fluctuation cores are spatially localized, do they correspond to the most fragile regions?
To test this, we independently perturbed each cell and measured the resulting recovery error (MSE of RGB channels within a local region; Figure~\ref{fig:vulnerability}).
Damage vulnerability was more strongly associated with mean state magnitude ($r = 0.639$) than with temporal fluctuation ($r = 0.325$), indicating that the primary predictor of fragility is how large a cell's internal state is rather than how much it fluctuates.
Here, mean state magnitude denotes the temporal mean of the Euclidean norm of the 16-dimensional cell state vector at each cell.
The spatial distribution of vulnerability (Figure~\ref{fig:vulnerability}B) was consistent with this asymmetry: highly vulnerable regions overlapped substantially with regions of large internal state magnitude, whereas the correspondence with fluctuation cores was weaker.
The limited contribution of fluctuation to vulnerability supports the interpretation that fluctuation cores primarily serve as sites of dynamical coordination rather than structural weak points.

Taken together, the preceding analyses establish that internal fluctuations in GNCA are intrinsic to the learned dynamics (not an artefact of stochastic cell updating) and persistent in the steady state.
They are spatially structured into channel-invariant cores whose locations do not simply coincide with either channel-specific mean-state patterns or regions of high damage vulnerability, suggesting that they reflect a distinct aspect of the system's dynamical organization. These cores are heterogeneously distributed across the cell population.
The following sections investigate their relationship to collective dynamics and functional role in self-repair.

\section{Collective Dynamics on a High-Dimensional Attracting State}
\label{sec:results_attractor}

Having established that fluctuations are structured, persistent, and spatially organized into cores, we next ask how the collective dynamics of the system are organized in the full high-dimensional state space.
\label{sec:results_umap}

To visualize the collective dynamical trajectory, we constructed a system-level feature vector at each time step by averaging the 16 channels at each alive cell to obtain a single scalar per cell, then concatenating these values across all ${\sim}500$ alive cells to yield a ${\sim}500$-dimensional representation. The resulting high-dimensional time series was embedded into two dimensions using UMAP \citep{mcinnes2018umap}.

\begin{figure}[!htb]
  \centering
  \includegraphics[width=\linewidth]{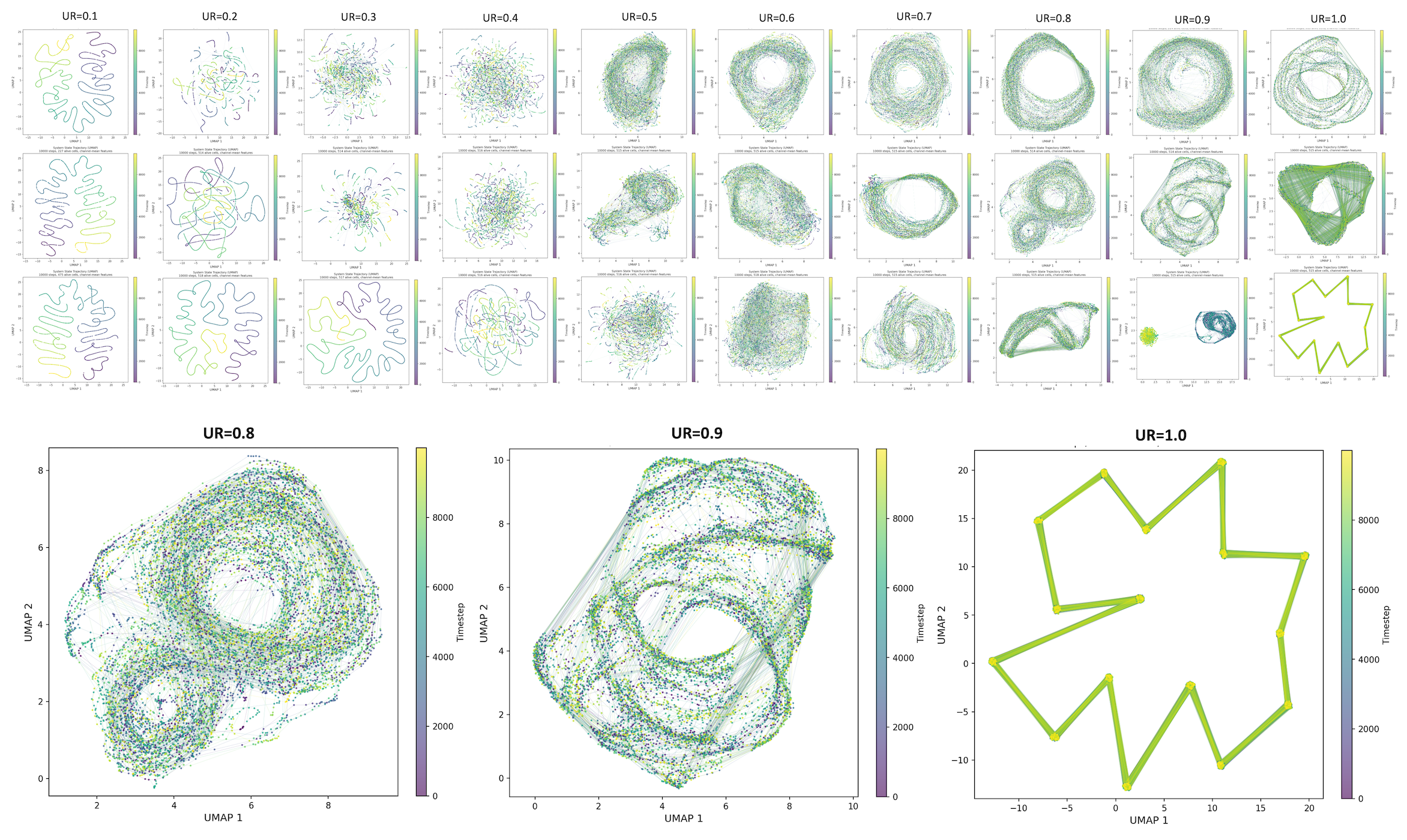}
  \caption{UMAP embedding of system-level state trajectories during undamaged steady-state operation. Top: overview across update rates UR $= 0.1$--$1.0$ (three samples each), showing a progressive transition from diffuse, random-walk-like trajectories at low UR to compact, structured paths at high UR. Bottom: enlarged views of representative trajectories at UR $= 0.8$, $0.9$, and $1.0$, illustrating compact recurrent structures, including loop-like paths and more complex folded trajectories. Color indicates time progression.\label{fig:umap_undamaged}}
\end{figure}

\begin{figure}[!htb]
  \centering
  \includegraphics[width=\linewidth]{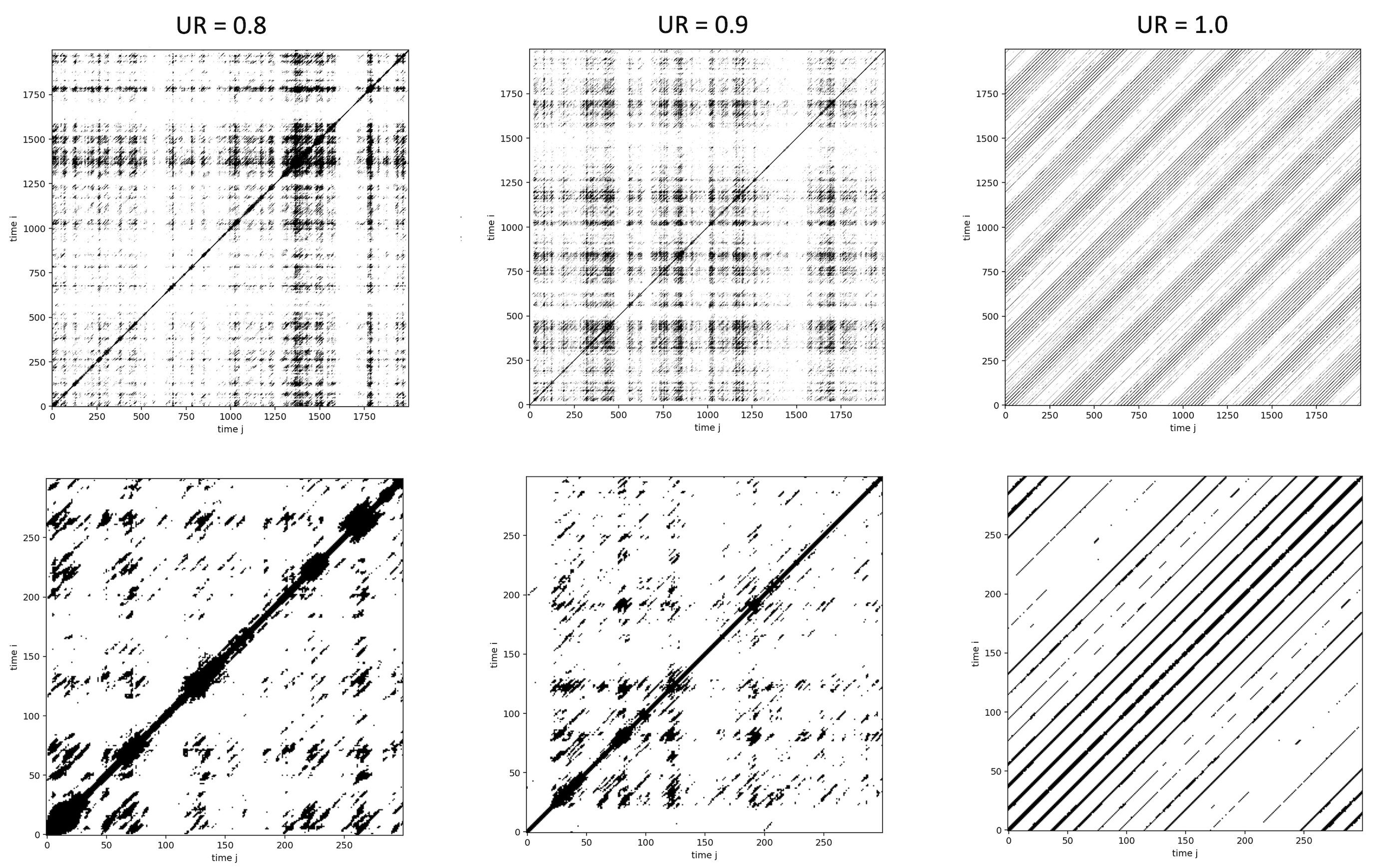}
\caption{
High-dimensional recurrence plots of system-level state trajectories during undamaged steady-state operation.
Recurrence plots were computed from z-scored high-dimensional state vectors by thresholding pairwise Euclidean distances at a fixed recurrence rate of $\mathrm{RR}=0.10$.
Top: full recurrence plots for representative trajectories at $\mathrm{UR}=0.8$, $0.9$, and $1.0$.
Bottom: enlarged views of the early-time windows.
Diagonal line segments indicate repeated visits to similar trajectory fragments in the original state space.
$\mathrm{UR}=0.8$ and $0.9$ show fragmented but structured recurrence, whereas $\mathrm{UR}=1.0$ shows regular diagonal bands consistent with a strong periodic component.
\label{fig:recurrnce_plot}}
\end{figure}

The embedding visualized low-dimensional manifold structure in the high-dimensional collective state space (Figure~\ref{fig:umap_undamaged}).
At sufficiently high update rates (UR~$\geq 0.5$), the trajectory traced out compact, structured paths in the embedding space, consistent with the steady-state dynamics being organized around an attracting state.
The specific geometry varied across trained models and update rates: some trajectories exhibited loop-like paths, while others showed more complex folded or recurrent structures. 
Supplementary analyses of additional target morphologies showed qualitatively similar compact and structured UMAP trajectories at high update rates, although the detailed trajectory geometry varied across target shapes and update rates (Appendix~\ref{app:analysis}).

To complement the UMAP visualization, we computed recurrence plots directly from the z-scored high-dimensional system-level state vectors (Figure~\ref{fig:recurrnce_plot}). Recurrence was defined by thresholding pairwise Euclidean distances at a fixed recurrence rate of $\mathrm{RR}=0.10$. The recurrence plots showed diagonal line segments, indicating repeated visits to similar trajectory fragments in the original state space rather than isolated state coincidences.

At $\mathrm{UR}=1.0$, where the dynamics are fully deterministic without stochastic cell updating, the recurrence plot exhibited regular diagonal bands, indicating a strong periodic component consistent with the oscillatory temporal autocorrelation observed in Section~3.2 and the loop-like UMAP trajectories. At $\mathrm{UR}=0.8$ and $0.9$, the recurrence structures were more fragmented but remained clearly organized, consistent with complex recurrent collective dynamics under stochastic updating.

However, even at $\mathrm{UR}=1.0$, some models exhibited more complex recurrent geometries comparable to those seen at $\mathrm{UR}=0.8$--$0.9$, suggesting that the learned update rule itself---independent of stochastic cell updating---can generate high-dimensional nonlinear recurrent dynamics.
At low update rates ($\mathrm{UR}\leq 0.3$), by contrast, the trajectories did not form compact structures but drifted continuously through state space, likely because stochastic cell updating dominates over the internal fluctuation dynamics and prevents the emergence of coherent recurrent organization.
This UR-dependent transition from diffuse trajectories to structured recurrent dynamics suggests that the balance between stochastic updating and learned dynamics plays a critical role in the emergence of collective organization.
Across the attracting recurrent state, the common feature is that the system does not converge to a fixed point but continuously traverses a structured region of state space, consistent with the persistence of fluctuations observed in the preceding analyses.

\begin{figure}[!htb]
  \centering
  \includegraphics[width=0.65\textwidth]{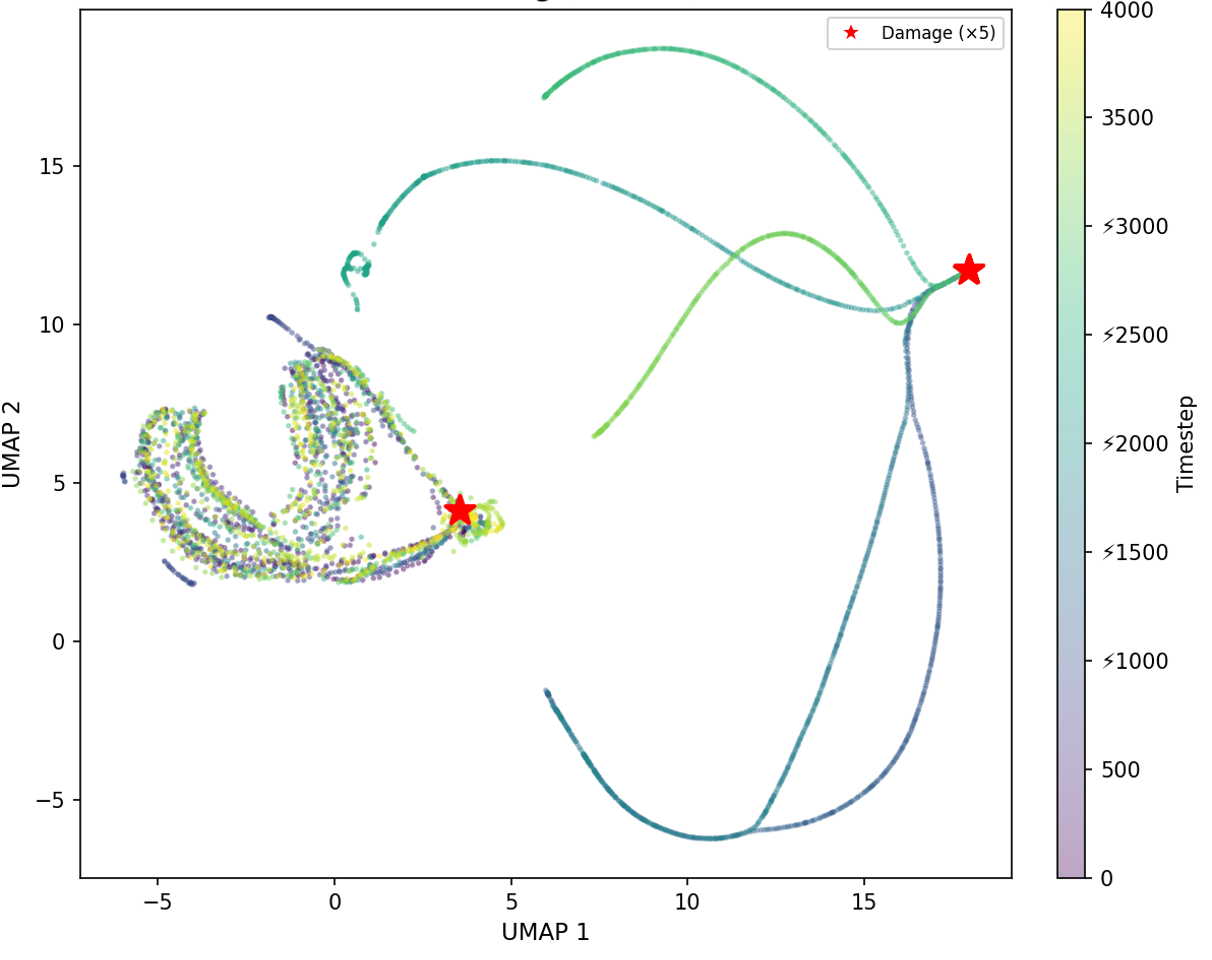}
  \caption{UMAP trajectory under periodic localized damage. Red stars indicate the time points of damage application. Each damage event displaces the system state to a remote region of latent space. The subsequent gradual re-convergence toward the resting recurrent region, extending over hundreds of steps, supports the interpretation of self-repair as return toward an attracting recurrent state through collective dynamical coordination.\label{fig:umap_damage}}
\end{figure}

When localized damage (radius 5 pixels, applied every 500 steps after 1000 relaxation steps) was applied periodically, the UMAP-embedded trajectory revealed dramatic dynamical signatures (Figure~\ref{fig:umap_damage}).
Each damage event caused the system state to jump to a remote region of the latent space, away from the resting recurrent region.
Following each perturbation, the trajectory exhibited a gradual return toward the original attracting state region, with the recovery path extending over hundreds of time steps.
This pattern is consistent with the interpretation of self-repair as a process of return toward an attracting recurrent region: damage displaces the system from its pre-damage recurrent trajectory, and the intrinsic dynamics of the learned update rule drive the system back toward the attracting recurrent region.
The extended time scale of recovery, relative to the rapid damage-induced displacement, suggests that the return trajectory involves coordinated multi-cell dynamics rather than independent local corrections.

\section{Local Damage, Global Response}
\label{sec:results_damage}

The return toward an attracting recurrent state observed in the preceding section raises the question of how localized damage affects the rest of the system and what dynamical ingredients are required for recovery.
We show that damage effects propagate far beyond the damage site and that recovery depends on micro-fluctuations across most of the organism.

\subsection{Nonlocal Propagation of Damage Effects}
\label{sec:results_nonlocal}

\begin{figure}[!htb]
  \centering
  \includegraphics[width=\linewidth]{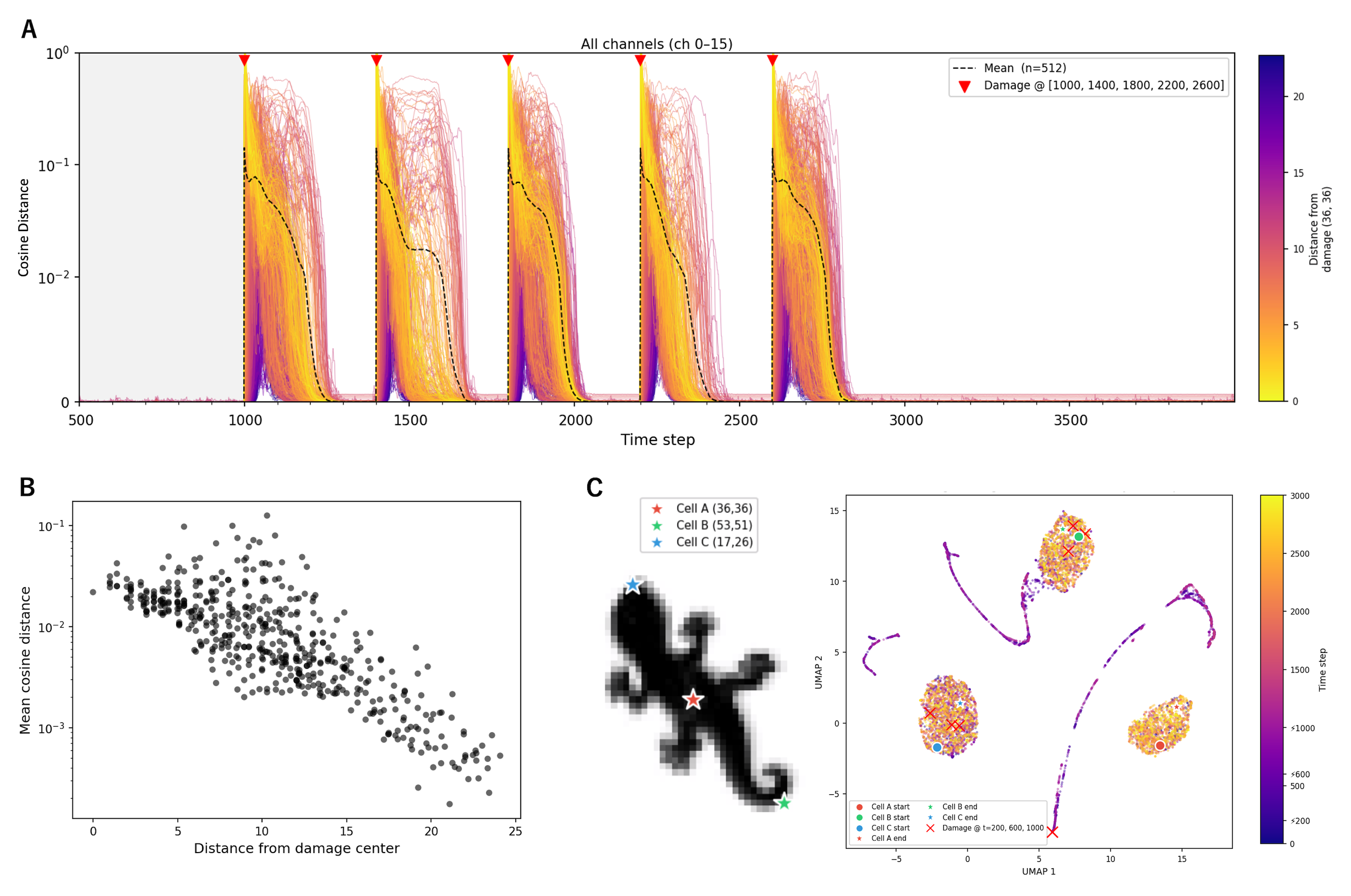}
  \caption{Nonlocal propagation of damage effects. (\textbf{A}) Spatial map of cosine distance from undamaged baseline state, defined as the temporal mean over the reference window, showing that cells far from the damage site also deviate from their resting state. (\textbf{B}) Mean cosine distance as a function of spatial distance from damage center, revealing a decay trend with notable mid-range excursions indicative of nonlocal recovery dynamics. (\textbf{C}) UMAP of individual cell state trajectories showing cell-level deviation and return, resembling the system-level return toward the recurrent state.\label{fig:cosine_distance}}
\end{figure}

Analysis of cell-level cosine distance from baseline following damage revealed that the effects of localized perturbation extended well beyond the immediate damage site (Figure~\ref{fig:cosine_distance}).
Cells at substantial distances from the damage center showed measurable deviations from their pre-damage baseline (measured as cosine distance from the temporal mean of undamaged states over steps 200--1000; Figure~\ref{fig:cosine_distance}A), demonstrating that damage is not a purely local event but propagates through the dynamical coupling of the system.
The mean cosine distance as a function of distance from the damage center (Figure~\ref{fig:cosine_distance}B) showed a general decay trend, as expected, but notably some cells at intermediate distances exhibited deviations comparable to or exceeding those of cells nearer the damage site.
Inspection of the time series showed that some intermediate-distance cells reached their peak deviation only after a delay relative to the initial damage event. We therefore interpret these delayed deviations as recovery-related responses, rather than as immediate direct propagation of the perturbation.
This suggests that they may reflect secondary dynamical responses recruited during the recovery process rather than direct perturbation propagation.
This nonlocal response pattern is consistent with the extended spatial correlations identified in \S\ref{sec:results_correlation} and suggests that the recovery process involves globally distributed dynamical adjustments, with distant cells actively contributing to coordinated repair.

UMAP embedding of individual cell trajectories (Figure~\ref{fig:cosine_distance}C) further supported this interpretation.
Individual cells showed recurrent trajectory structure, with damage causing a deviation from the pre-damage trajectory followed by a return trajectory.
This suggests that the system-level recovery process is accompanied by analogous perturbation-and-return dynamics at the single-cell level, indicating a nested organization of recovery dynamics.

\begin{figure}[!htb]
  \centering
  \includegraphics[width=\linewidth]{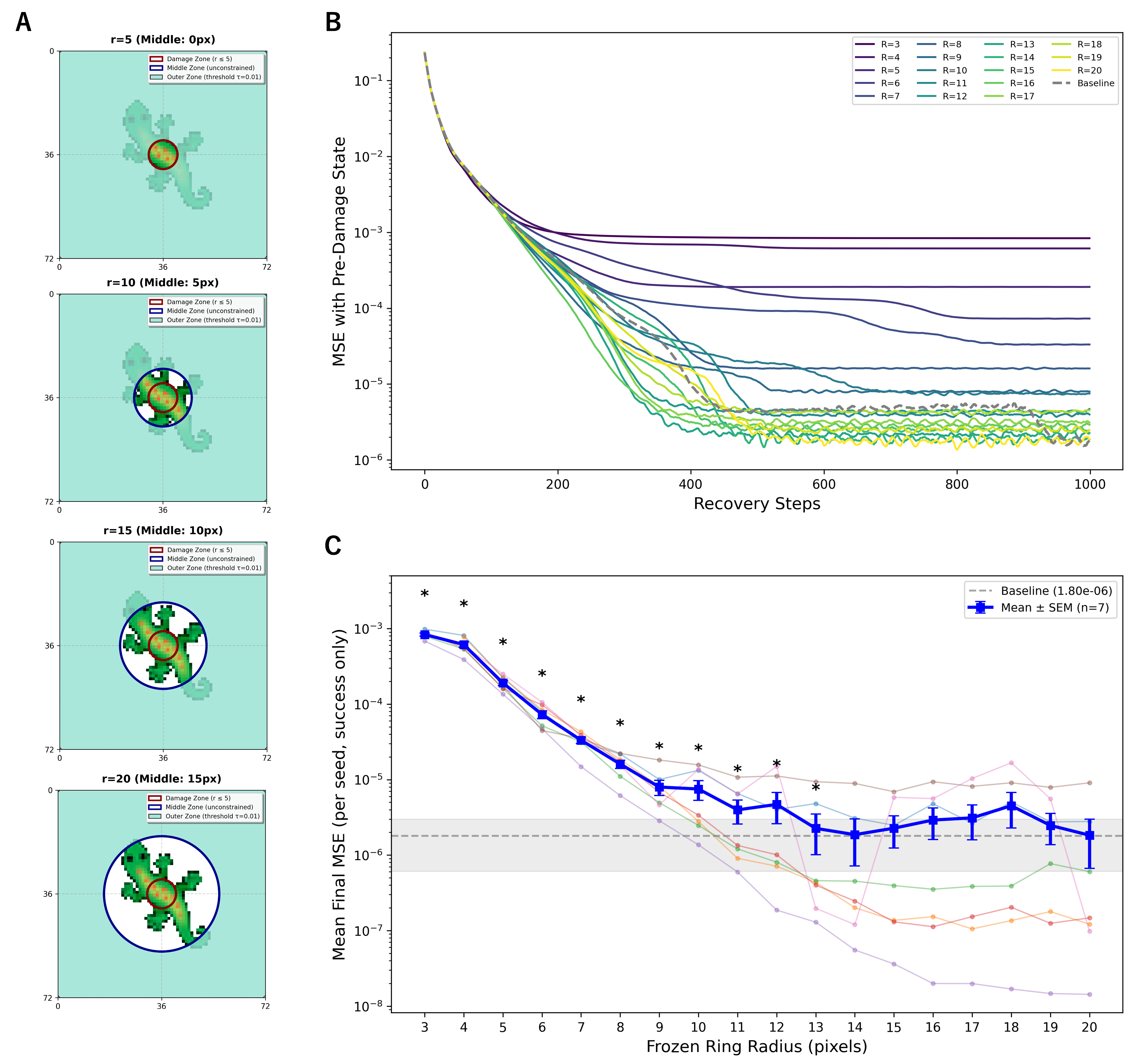}
  \caption{Fluctuation contribution radius experiment. (\textbf{A}) Schematic of the experimental design: within a permissive zone of radius $r$ centered on the damage site, all updates proceed normally; outside, updates with mean magnitude $\leq 0.01$ are blocked while larger changes propagate freely. The radius $r$ is swept from 5 to 20 pixels. (\textbf{B}) Recovery dynamics (MSE of RGB channels within the damage region) over recovery steps for each permissive radius. Smaller radii show persistently elevated MSE, indicating impaired recovery. (\textbf{C}) Final MSE as a function of permissive radius. Recovery quality degrades significantly for permissive radii up to $r = 13$, with the effect vanishing at $r = 14$ (${\approx}74\%$ of alive cells). This indicates that effective recovery depends on distributed small-magnitude updates across the majority of the organism.\label{fig:radius}}
\end{figure}

\subsection{Organism-Wide Micro-Fluctuation Activity Supports Recovery}
\label{sec:results_radius}

To directly test whether the distributed small-magnitude update associated with micro-fluctuations contributes to recovery, we designed an intervention experiment.
A circular permissive zone of radius $r$ was defined around the damage site (damage radius = 5 pixels).
Within this zone, all updates proceeded without restriction.
Outside the zone, any cell whose update had a mean absolute magnitude across channels of 0.01 or less was prevented from updating at that step, effectively suppressing micro-fluctuations while still allowing larger state changes to propagate.
This threshold of 0.01 was chosen based on the activity distributions characterized in \S\ref{sec:results_stats}: the CCDF analysis (Figure~\ref{fig:basic_characters}C) shows that the vast majority of per-cell activity in the undamaged steady state falls at or below this magnitude, whereas damage-induced activity frequently exceeds it.
The threshold thus selectively blocks the baseline micro-fluctuations that constitute normal resting dynamics while permitting the larger state changes associated with active repair signals to pass through.
The permissive radius was swept from $r = 5$ to $r = 20$ pixels at UR~$= 0.8$ (Figure~\ref{fig:radius}A), with 10 trials per model across all 8 models.
Recovery quality was measured as the MSE of RGB channels within the damage region.

The recovery trajectories (Figure~\ref{fig:radius}B) showed that smaller permissive radii led to persistently elevated MSE, indicating sustained impairment of the repair process.
Recovery quality was significantly impaired for permissive radii up to $r = 13$ pixels, with significance vanishing at $r = 14$ (Mann-Whitney $U$ test against the unsuppressed baseline for each radius, $n = 8$; Figure~\ref{fig:radius}C).
A permissive zone of radius 14 encompasses approximately 74\% of all alive cells, indicating that effective recovery depends on micro-fluctuation-related activity across much of the organism, well beyond the immediate vicinity of the damage site.

This result is important for two reasons.
First, it indicates that micro-fluctuation-related updates are not epiphenomenal: suppressing distributed small-magnitude updates impairs the system's ability to recover from damage.
Second, the extensive spatial range of the contribution zone---encompassing the majority of the organism---demonstrates that recovery is a fundamentally organism-wide process, not a local repair confined to the damage vicinity. This is consistent with the extended spatial correlations measured in \S\ref{sec:results_correlation} and reinforces the view that fluctuation-mediated coordination operates at the scale of the entire system.

In summary, localized damage triggers organism-wide dynamical responses, and recovery depends on micro-fluctuation-related activity across much of the cell population, underscoring the collective nature of the repair process. This suggests that micro-fluctuation-related small updates contribute to the coordination of repair beyond the immediate damage site.
The next section investigates the information-theoretic mechanisms that mediate this coordination.

\section{Information-Theoretic Signatures of Maintenance and Repair}
\label{sec:results_information}

The preceding sections established that fluctuations are structured and that distributed small-magnitude updates associated with baseline fluctuation dynamics contribute to recovery, and that damage recovery involves globally distributed dynamics.
We now ask how information flows through the system during maintenance and repair, and what computational regime underlies each state.
We address these questions using transfer entropy to characterize directed information flow, and information decomposition methods---integrated information decomposition ($\Phi$ID) for inter-channel temporal coupling and partial information decomposition (PID) for spatial computation---to reveal the structure of multi-source and multi-channel information processing.

\subsection{Spatially Differentiated Information Flow During Repair}
\label{sec:results_te}

To characterize directed information flow, we estimated transfer entropy (TE) \citep{schreiber2000measuring} between pairs of neighboring cells.
Each cell's 16-channel time series was discretized into 8 bins via quantile binning, and TE was computed using a history length of $k = 1$ via the \texttt{pyinform} library.  Bin edges were fitted independently for each scalar time series within each condition and analysis window, without pooling across cells, channels, spatial locations, or conditions.
The same per-series quantile discretization procedure was used for the $\Phi$ID and PID analyses below.

For each cell, we computed both outgoing TE (cell $\to$ neighbor) and incoming TE (neighbor $\to$ cell) across all 8 spatial neighbors and 16 channels.
To construct a spatial vector field, we computed the \textit{asymmetry vector} at each cell: for each neighbor at relative displacement $(\Delta x, \Delta y)$, the net information asymmetry $(\text{TE}_\text{out} - \text{TE}_\text{in})$ was weighted by the displacement direction, summed across neighbors, and normalized to unit length.
Although 8-bin quantile discretization was used for the main analysis, the qualitative structure of the resulting TE asymmetry field was preserved under 4-, 8-, and 16-bin discretizations (Appendix~\ref{app:analysis}).

Damage (radius 5 pixels) was applied every 500 steps, with 1000 relaxation steps before the first cycle and 100 cycles in total.
We focused on the early post-damage window (steps 0--19 immediately after each damage event), concatenating 100 cycles to yield 2000 time steps for TE estimation.

\begin{figure}[!htb]
  \centering
  \includegraphics[width=\linewidth]{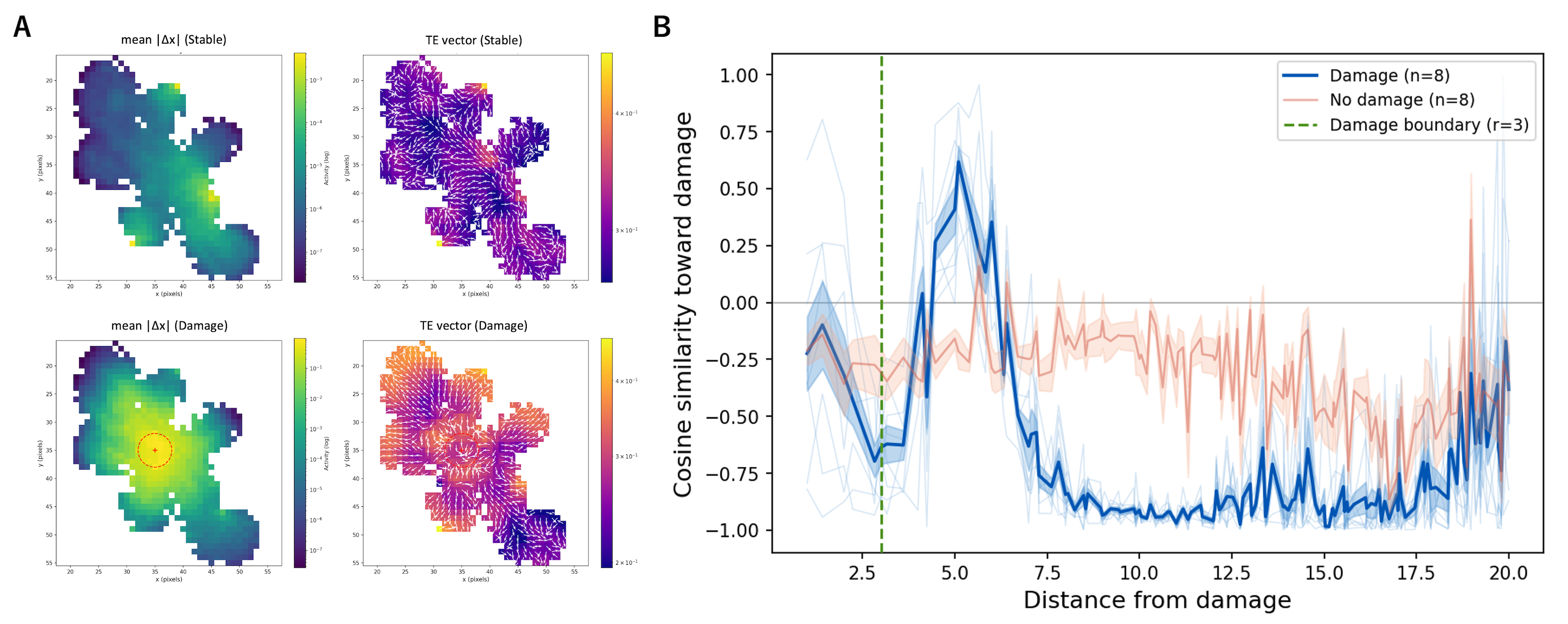}
  \caption{(\textbf{A}) Transfer entropy analysis comparing undamaged steady state (top) and early post-damage response (bottom). Left column: mean activity $|\Delta\mathbf{s}|$; right column: TE asymmetry vector field. In the undamaged state, the TE asymmetry field points outward from fluctuation cores. After damage, a ring of elevated TE forms around the damage site, with inward flow within the ring and outward flow beyond it.
  (\textbf{B}) Mean cosine similarity between each cell's TE asymmetry vector and the unit vector pointing toward the damage site, as a function of distance from the damage center ($n = 8$, mean $\pm$ individual seeds). Positive values indicate inward (repair-directed) flow; negative values indicate outward (propagation) flow. A sharp transition from inward to outward flow occurs just beyond the damage boundary (dashed line), revealing the spatial boundary between corrective and propagative information dynamics.\label{fig:te_fields}}
\end{figure}

Computing these fields under two conditions---undamaged steady state and early post-damage---revealed a fundamental reconfiguration of information flow during repair (Figure~\ref{fig:te_fields}A).
In the undamaged condition, the regions of elevated TE magnitude tended to overlap with the fluctuation cores identified in \S\ref{sec:results_cores}, and the TE vector field exhibited a predominantly outward flow pattern emanating from these regions.
This suggests that fluctuation cores serve as sources of information that is broadcast to the surrounding tissue during normal maintenance.

In the early damage condition, the flow pattern reorganized dramatically.
A ring-shaped region of elevated TE magnitude formed around the damage site, marking a boundary between two distinct information flow regimes.
Within this ring, asymmetry vectors pointed predominantly inward toward the damage center, consistent with an immediate corrective response.
Beyond the ring, the flow direction reversed to outward, indicating simultaneous propagation of perturbation-related signals to more distal regions.
This spatial coexistence of repair-directed and propagation-directed information flow within the same time window suggests that the system initiates corrective action locally while broadcasting damage information globally.

To quantify this spatial structure, we computed the cosine similarity between each cell's asymmetry vector and the unit vector pointing toward the damage site, and plotted the mean as a function of distance from the damage center across 8 models (Figure~\ref{fig:te_fields}B).
The profile showed a sharp peak in inward flow (positive cosine similarity) just beyond the damage boundary, followed by a rapid transition to outward flow (negative cosine similarity) at slightly greater distances.
In contrast, the undamaged condition showed near-uniform outward flow across all distances.
This spatial profile confirms that the post-damage response is spatially differentiated---corrective inward flow is concentrated near the damage site while perturbation-related information is simultaneously broadcast outward to distal regions.
Because neighboring cells share overlapping perception neighborhoods, these pairwise TE estimates should be read as directed predictive asymmetries rather than as direct causal transmission; common-driver effects cannot be fully excluded, although the asymmetry measure ($\text{TE}_\text{out} - \text{TE}_\text{in}$) partially cancels symmetric common-driver contributions.

\subsection{Inter-Channel Differentiation and Synergistic Integration}
\label{sec:results_interchannel}

To characterize how information is distributed across channels, we applied integrated information decomposition ($\Phi$ID) \citep{mediano2025integrated} (see Appendix~\ref{app:metrics} for formal definitions).
Whereas standard partial information decomposition (PID) decomposes the information that multiple \textit{source} variables carry about a single \textit{target} variable, $\Phi$ID extends this framework to bivariate temporal processes: given two time series $X$ and $Y$, it considers the four variables $(X_t, X_{t+\tau}, Y_t, Y_{t+\tau})$ and decomposes the temporal mutual information into 16 atoms capturing how information is shared, transferred, and created across the past and future of each process.
$\Phi$ID is therefore well suited to characterizing the temporal coupling between channel pairs at a given cell, where the question is not ``how do multiple spatial inputs inform a target'' but ``how do two channels exchange and generate information over time.''
We used the \texttt{phyid} library with minimum mutual information (MMI) redundancy.
We report three key diagonal atoms: redundancy, unique information, and synergy. 

\begin{figure}[!htb]
  \centering
  \includegraphics[width=\linewidth]{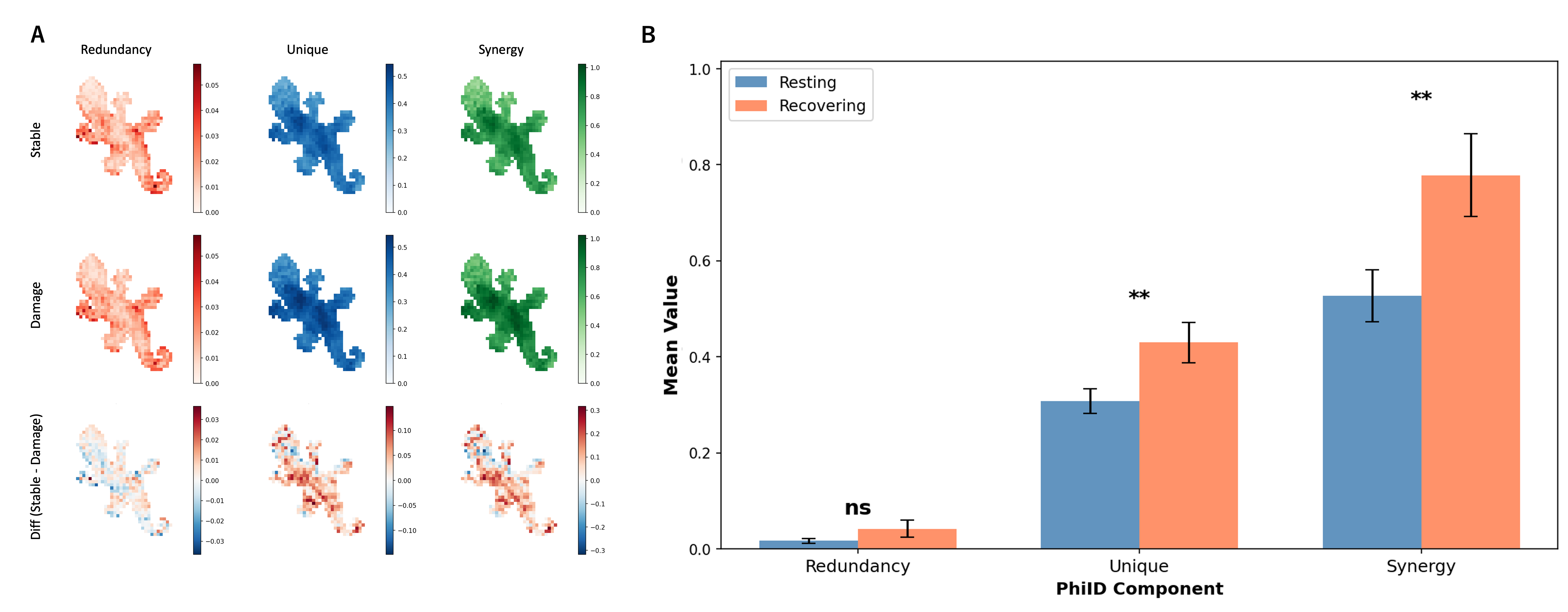}
  \caption{Inter-channel $\Phi$ID across all channel pairs under undamaged and damaged conditions. Synergy (\texttt{sts}) is the dominant component, unique information is substantial, and redundancy (\texttt{rtr}) is negligible---indicating a computational substrate that combines integrative and specialized processing without redundant duplication. Damage increases all components, suggesting intensified inter-channel coupling during repair. Error bars indicate SEM across independently trained models initialized with different random seeds.
Values were first averaged over alive cells and channel pairs within each model, and statistical comparisons were performed across models using Wilcoxon signed-rank tests. Asterisks indicate significant resting--recovery differences ($^{**}p<0.01$). \label{fig:interchannel_pid}}
\end{figure}

$\Phi$ID was computed for all channel pairs $(c_i, c_j)$ at each alive cell.
All continuous variables were discretized into 8 bins via quantile binning with a time lag of $\tau = 1$.
$\Phi$ID values were first averaged over channel pairs and alive cells within each independently trained model.
Undamaged and damaged conditions were then compared across independently trained models using Wilcoxon signed-rank tests.

This analysis revealed a distinctive information-theoretic organization of multi-channel computation in GNCA (Figure~\ref{fig:interchannel_pid}).
Synergy was the largest component, indicating that a substantial portion of the information about each channel's future state arises from the \textit{joint} configuration of channel pairs rather than from either channel alone.
Unique information was also appreciable---approximately half the magnitude of synergy---demonstrating that individual channels additionally carry independent predictive information about their own future dynamics.
By contrast, redundancy was an order of magnitude smaller than the other components, indicating that channel pairs share very little overlapping information.
This pattern---high synergy, substantial unique information, and negligible redundancy---suggests that the 16 channels operate as a computational substrate combining both integrative (synergistic) and specialized (unique) information processing, while avoiding redundant duplication.

Under the damaged condition, all $\Phi$ID components increased relative to the undamaged baseline, with a particularly notable increase in total information.
This suggests that damage induces a global intensification of inter-channel information coupling, consistent with the system entering a more informationally active state during repair.
The preservation of the overall dominance structure under damage indicates that the fundamental computational architecture is maintained during recovery, even as the overall level of information exchange increases.

\subsection{Synergy-to-Redundancy Regime Shift in Spatial Computation}
\label{sec:results_spatial_pid}

\begin{figure}[!htb]
  \centering
  \includegraphics[width=\linewidth]{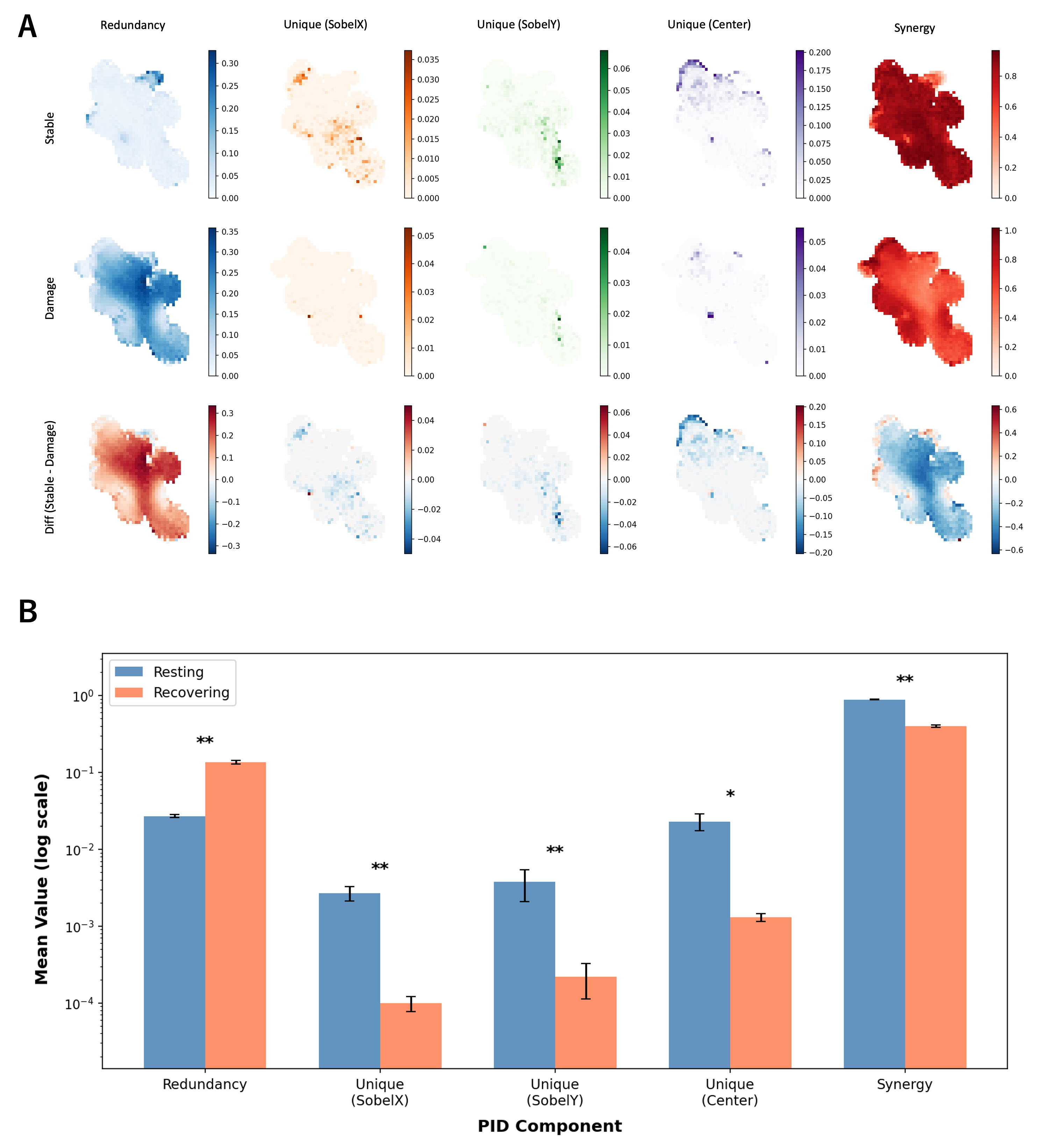}
  \caption{Spatial partial information decomposition comparing resting and recovery conditions. 
  During resting self-maintenance, synergy dominates, indicating integrative spatial computation. 
  During recovery, redundancy increases with corresponding decreases in unique information and synergy, consistent with a shift toward redundancy-enhanced coordination during repair.
PID components were first averaged over alive cells within each independently trained model, and statistical comparisons were performed across independently trained models using Wilcoxon signed-rank tests.
Error bars indicate SEM across independently trained models.
Asterisks indicate significant resting--recovery differences ($^{*}p<0.05$, $^{**}p<0.01$).
  \label{fig:spatial_pid}}
\end{figure}

While $\Phi$ID characterizes temporal coupling between channel pairs at the same cell, a complementary question is how \textit{spatial} information sources combine to determine each cell's update.
For this, we used standard PID \citep{williams2010nonnegative} (see Appendix~\ref{app:metrics}), which decomposes the information that multiple source variables carry about a single target---a natural fit for the multi-input spatial computation of the GNCA update rule.
Specifically, we applied the Williams-Beer estimator (\texttt{PID\_WB} from the \texttt{dit} library) with three source variables at time $t$---the horizontal Sobel gradient (SobelX), the vertical Sobel gradient (SobelY), and the center cell state---and target $\Delta s(t{+}1) = s(t{+}1) - s(t)$.
The Sobel gradients were computed using standard $3 \times 3$ kernels matching the GNCA perception module.
For the three-source case, the Williams-Beer PID lattice yields a richer decomposition than the two-source case; we report the full three-way redundancy $R(S_1, S_2, S_3; T)$, individual unique information terms, and the full three-way synergy $S(S_1, S_2, S_3; T)$.
We compared resting versus recovery conditions.
For statistical comparison, PID components were first averaged over alive cells within each independently trained model. Resting and recovery conditions were then compared across independently trained models using Wilcoxon signed-rank tests. 

The decomposition revealed a pronounced regime shift (Figure~\ref{fig:spatial_pid}).
During the resting state, synergy was the dominant component, indicating that predictive information about local state dynamics was primarily carried by the joint, nonlinear combination of spatial gradient and center state features rather than by any single feature alone.
This synergy-dominant regime suggests that the system performs integrative spatial computation during self-maintenance, combining multiple sources of contextual information to determine each cell's update.

During recovery from damage, the information structure shifted significantly.
Redundancy increased substantially, while both unique information and synergy decreased relative to the resting baseline.
The increase in redundancy and the decreases in unique information and synergy were statistically significant across independently trained models (Wilcoxon signed-rank tests).
The increase in redundancy during recovery is consistent with the hypothesis that the system transitions to a more coordinated, error-correcting computational mode in which multiple spatial features carry overlapping information about the required state correction.
Redundancy-enhanced computation may provide robustness by ensuring that corrective signals are transmitted through multiple independent channels, reducing the vulnerability of the repair process to local errors or signal degradation.

Overall, the TE, $\Phi$ID, and PID analyses reveal a spatially and functionally differentiated information structure during maintenance and repair. 
Supplementary analyses using additional target morphologies showed qualitatively similar trends in these information-theoretic signatures, including damage-induced reorganization of TE asymmetry fields, intensified inter-channel coupling, and increased redundancy during recovery (Appendix~\ref{app:analysis}). 
These results suggest that the observed information structure is not specific to a single target morphology, but is qualitatively reproduced across the additional morphologies tested under the same architecture and training regime.

\section{Discussion}
\label{sec:discussion}

We demonstrated that internal fluctuations in GNCA are not residual noise but spatially structured, dynamically organized components associated with distributed small-magnitude updates whose suppression impairs self-repair. 
We further showed that the collective dynamics are organized around attracting recurrent organization, and that damage recovery can be interpreted as a return toward a pre-damage recurrent state mediated by spatially differentiated information flow and a shift from synergy-dominant to redundancy-enhanced computation.
We discuss the implications of these findings below.

\paragraph{Fluctuations as dynamical and informational infrastructure.}
The structured fluctuations we observe---intrinsic to the learned dynamics, spatially organized into channel-invariant cores, and robust across update mechanisms---parallel the functional role of fluctuations in biological systems.
In gene regulatory networks, molecular noise supports cellular decision-making \citep{eldar2010functional,tsimring2014noise} and cell-to-cell variability reflects the multi-stable attractor structure of the underlying dynamics \citep{huang2009heterogeneity}.
The functional role of these fluctuations is supported by an intervention experiment in which suppressing distributed small-magnitude updates outside a permissive radius significantly impairs regeneration, indicating that baseline fluctuation dynamics are not merely correlated with but contribute substantially to self-repair.
More broadly, our findings instantiate Kaneko and Ikegami's concept of homeochaos \citep{kaneko1992homeochaos}: dynamic stability sustained through---rather than despite---high-dimensional structured variability.
GNCA can be viewed as a neural counterpart of coupled map lattices (CML) \citep{kaneko1992overview,kaneko2001complex}, in which local nonlinear maps are replaced by learned neural update rules.
Indeed, the spatial correlation scaling and the attracting recurrent state we observe are consistent with CML phenomenology, suggesting that GNCA---despite being trained rather than hand-designed---converge on dynamical regimes studied in the physics of spatially extended nonlinear systems \citep{nicolis1977selforg}.
%
%

\paragraph{Collective dynamical homeostasis through organizing attracting states.}
Framing self-repair as a return to an attracting state connects GNCA to the broader theory of attractor dynamics in complex systems, where robustness arises from the basin structure of the dynamical landscape rather than from explicit error-correction mechanisms \citep{huang2005cellfates,waddington1957strategy}.
The nested attracting-state structure we observe---at both the system and single-cell levels---resonates with the ultrastability concept of \citet{ashby1960design} and the adaptivity framework of \citet{dipaolo2005autopoiesis}, in which adaptive systems maintain global homeostasis through hierarchically organized feedback.
This perspective may also explain why even GNCA models trained without damage exhibit resilience to perturbation \citep{mordvintsev2020growing}.
If learning to maintain a target morphology inherently gives rise to an attracting recurrent state, then some damage tolerance follows as a natural consequence of the surrounding basin structure, regardless of whether the model was explicitly trained to recover.
Damage-inclusive training would then serve to widen the basin of attraction rather than to create recovery capacity \textit{de novo}.
However, the dynamical-systems perspective developed above characterizes that the system returns to its attracting state, but does not reveal how the return is internally organized or where different functional roles are spatially located.
The information-theoretic analyses developed below address this gap.


\paragraph{Spatially organized information structure in maintenance and repair.}
Even before damage, the system exhibits a pre-existing spatial information structure.
Transfer entropy indicates that fluctuation cores are associated with sustained outward predictive asymmetry, consistent with their acting as information sources, a continuous-field analogue of the coherent structures identified in classical CA \citep{lizier2008local}.
Spatial PID, extending earlier spatiotemporal filter analyses of classical CA \citep{flecker2011pid,finn2018pointwise} to learned continuous-state systems, further shows that this resting computation is synergy-dominant: each cell's update arises from integrative combination of spatial features, actively maintained by ongoing fluctuations rather than assembled \textit{de novo} upon damage.

Damage reorganizes this information structure without dismantling it.
Transfer entropy shows a spatially differentiated response: corrective inward flow near the damage site coexists with sustained outward broadcasting in distal regions, suggesting that distinct spatial zones perform distinct informational functions simultaneously.
Inter-channel $\Phi$ID shows intensified coupling across all channel pairs while preserving the synergy $>$ unique $>$ redundancy hierarchy, and spatial PID reveals a shift toward redundancy-enhanced computation, echoing synergy-redundancy transitions observed between cognitive states in neuroscience \citep{luppi2024information}.
Together, these results suggest that self-repair is not a single dynamical process but involves a spatial division of informational labor---with fluctuation cores as information sources, damage-proximal cells as corrective sinks, and distal cells as broadcasting relays---that self-organizes from the same local update rule that produces the resting dynamics.

Notably, this collective attracting state emerges from a population of cells governed by an identical update rule.
During development, spatial context drives spontaneous differentiation into distinct state configurations---as evidenced by the strong channel-specific spatial patterns in Figure~\ref{fig:spatial_maps}A---while the micro-fluctuations of these differentiated cells collectively generate the system-level attracting state dynamics that sustain homeostasis.
This process is analogous to isologous diversification in coupled dynamical systems, where identical units spontaneously break symmetry to produce a functionally differentiated population whose collective dynamics are richer than those of any individual \citep{kaneko1997isologous,furusawa2012stemcell}.
The mutual dependence between differentiation and collective stability constitutes a form of collective homeostasis: cells require the collective context to maintain their differentiated states, while the collective attracting state depends on the coordinated contributions of differentiated cells.
In this view, individuality is not a precondition for collective organization but rather its product: the community of interacting cells comes first, and functional differentiation emerges through the dynamics of that community.

\paragraph{Precarious self-maintenance.}
A useful way to frame the GNCA dynamics is not as static stability, but as precarious self-maintenance: persistence depends on the continued operation of the very processes that constitute the maintained organization. This notion is closely related to Di Paolo's account of \emph{precariousness} \citep{dipaolo2005autopoiesis}, in which a system's organization persists only through ongoing activity whose interruption would undermine that organization. 

Our analyses make this concrete. The resting state is not a fixed point but an attracting recurrent state, so the morphology is actively maintained rather than passively stable. Correspondingly, suppressing micro-fluctuation-related small updates impairs recovery: when this constitutive activity is curtailed, the organization becomes less able to restore itself after perturbation. This provides an empirical analogue of precariousness, in which persistence depends on the ongoing activity of the system's own constitutive dynamics. 

From this perspective, maintenance and repair are not separate mechanisms but two expressions of the same precarious organization. The distributed fluctuation dynamics that sustain the morphology during undamaged operation are reorganized during recovery, supporting return toward the pre-damage recurrent state and system-wide coordination after perturbation. GNCA thus provide a tractable artificial model of collective homeostasis in which robustness arises from continuously active, precarious self-maintenance rather than static equilibrium.

\paragraph{Limitations and future work.}
\label{sec:limitations}

Several limitations of the present study point toward productive directions for future research.
Our information-theoretic estimates (TE, $\Phi$ID, PID) depend on discretization and parameter choices, and the attracting recurrent dynamics identified here remain suggestive rather than a formal dynamical classification; additional diagnostics such as Lyapunov spectra, fractal-dimension estimates, or recurrence quantification analysis would strengthen the dynamical interpretation.

While the main analyses focus on the lizard morphology, supplementary analyses of additional target morphologies showed qualitatively similar fluctuation-core patterns, recurrent trajectories, TE responses, and information-decomposition signatures.
A more systematic comparison across architectures, perception filters, hidden-channel capacities, alive-mask definitions, and boundary conditions remains future work; our robustness claims are therefore limited to qualitative generality across the tested morphologies under a single architecture and training regime.


Our analyses focus exclusively on the post-developmental regime---the dynamics of the system after it has reached the target morphology.
The morphogenetic growth process itself, from a single seed cell to the full pattern, likely involves qualitatively different dynamical regimes (e.g., symmetry-breaking transitions, progressive spatial differentiation) that may exhibit distinct information-theoretic signatures.
Characterizing the developmental trajectory and its relationship to the steady-state attracting structure identified here remains an important open question.


Finally, the damage protocol studied here---periodic localized zeroing---represents one specific perturbation type.
Investigating distributed noise, partial channel corruption, or multi-site damage would clarify the generality of the spatially differentiated repair dynamics and the information regime shift we observed.

\section{Conclusions}
\label{sec:conclusion}

This study provides a comprehensive empirical characterization of the internal dynamics underlying self-maintenance and self-repair in Growing Neural Cellular Automata.
Our central finding is that internal fluctuations are not residual noise but structured, spatially organized dynamical components whose suppression impairs recovery from damage.

The collective dynamics of the GNCA system exhibit attracting recurrent organization, and self-repair corresponds to return toward a pre-damage recurrent state following perturbation-induced displacement.
Suppressing distributed small-magnitude updates outside a permissive radius encompassing approximately 74\% of alive cells significantly impairs recovery, suggesting that micro-fluctuation-related update activity contributes to repair at a near-global scale.
This process is mediated by a spatially differentiated information flow pattern---revealed by transfer entropy---in which corrective inward flow near the damage site coexists with outward perturbation propagation at greater distances.
The information-theoretic regime shifts from synergy-dominant computation during resting maintenance to redundancy-enhanced coordination during recovery.
This suggests an adaptive computational strategy that balances efficiency and robustness.

These findings contribute to a dynamical systems perspective on self-organization in neural cellular automata and suggest that the principles of attracting dynamics, structured fluctuations, and information regime shifts may provide a general framework for understanding robust self-repair in decentralized computational systems.

\vspace{6pt}


\section*{Acknowledgments}

\paragraph{Author Contributions.}
Conceptualization, A.M.; methodology, A.M.; software, A.M.; formal analysis, A.M.; investigation, A.M.; writing---original draft preparation, A.M.; writing---review and editing, A.M., H.S. and T.I.; visualization, A.M.; supervision, T.I. All authors have read and agreed to the published version of the manuscript.

\paragraph{Funding.}
This work was partially supported by JSPS KAKENHI Grant Number 23K16982, 24H00707.

\paragraph{Data Availability.}
The code and trained model checkpoints used in this study are available on a dedicated GitHub repository upon request to Atsushi Masumori (atsushi.masumori@gmail.com).

\appendix
\section{Metric Definitions}
\label{app:metrics}

\subsection{Cosine Distance from Baseline}
For a cell at position $(i,j)$ with state vector $\mathbf{s}_{i,j}(t)$ at time $t$ and baseline state $\bar{\mathbf{s}}_{i,j}$ (temporal mean during an undamaged reference period), the cosine distance is defined as:

\begin{equation}
  d_{i,j}(t) = 1 - \frac{\mathbf{s}_{i,j}(t) \cdot \bar{\mathbf{s}}_{i,j}}{\|\mathbf{s}_{i,j}(t)\| \, \|\bar{\mathbf{s}}_{i,j}\|} .
\end{equation}

Values range from 0 (identical direction) to 2 (opposite direction), with 1 indicating orthogonality.

\subsection{Transfer Entropy and Asymmetry Vector Field}
Transfer entropy from source cell $X$ to target cell $Y$ is defined as \citep{schreiber2000measuring}:

\begin{equation}
  \text{TE}_{X \to Y} = \sum p(y_{t+1}, y_t^{(k)}, x_t^{(k)}) \log \frac{p(y_{t+1} \mid y_t^{(k)}, x_t^{(k)})}{p(y_{t+1} \mid y_t^{(k)})},
\end{equation}

with history length $k = 1$ and 8-bin quantile discretization. We use k = 1 because the GNCA update rule is a one-step local rule, so k = 1 captures direct one-step predictive dependencies; longer-lag analysis is left to future work. 
For each cell $Y$, we compute both outgoing $\text{TE}_{Y \to N}$ and incoming $\text{TE}_{N \to Y}$ for each neighbor $N$ at displacement $\boldsymbol{\delta}_N$.
The asymmetry vector at cell $Y$ is the normalized sum of directionally weighted net information flow:

\begin{equation}
  \mathbf{v}_Y = \hat{\mathbf{u}}, \quad \mathbf{u} = \sum_{N \in \mathcal{N}(Y)} \left( \text{TE}_{Y \to N} - \text{TE}_{N \to Y} \right) \boldsymbol{\delta}_N,
\end{equation}

where $\hat{\mathbf{u}} = \mathbf{u} / \|\mathbf{u}\|$.
This vector points in the direction of net information outflow from cell $Y$.

\subsection{Integrated Information Decomposition}
The inter-channel analysis uses integrated information decomposition ($\Phi$ID) \citep{mediano2025integrated}, which extends the PID lattice to the temporal domain.
For each channel pair $(c_i, c_j)$ at a given cell, $\Phi$ID considers four variables: the past and future of each channel ($c_i^{t}, c_i^{t+\tau}, c_j^{t}, c_j^{t+\tau}$) and decomposes the mutual information into 16 atoms.
We report the key diagonal atoms: redundancy (\texttt{rtr}), unique information ((\texttt{xtx}~$+$~\texttt{yty})/2), and synergy (\texttt{sts}), computed with minimum mutual information (MMI) redundancy.
All time series were discretized into 8 bins via quantile binning with $\tau = 1$.

\subsection{Partial Information Decomposition}
For the spatial analysis, we used the partial information decomposition (PID) framework \citep{williams2010nonnegative} with the Williams-Beer estimator.
PID decomposes the mutual information $I(\{S_1, \ldots, S_n\}; T)$ that a set of source variables carries about a target $T$ into non-negative atoms corresponding to redundancy, unique information, and synergy.

For two sources, the decomposition is:
\begin{equation}
  I(S_1, S_2; T) = \underbrace{I_\cap(S_1; S_2; T)}_{\text{redundancy}} + \underbrace{U(S_1; T)}_{\text{unique}_1} + \underbrace{U(S_2; T)}_{\text{unique}_2} + \underbrace{CI(S_1; S_2; T)}_{\text{synergy}},
\end{equation}
where redundancy $I_\cap$ is the information that both sources provide about $T$, unique information $U$ is the information carried exclusively by one source, and synergy $CI$ is the information available only from the joint observation of both sources.
The Williams-Beer measure defines redundancy via the minimum specific information:
\begin{equation}
  I_\cap(S_1; S_2; T) = \sum_{t} p(t) \min_{i} I_{\text{spec}}(S_i; T{=}t),
\end{equation}
where $I_{\text{spec}}(S_i; T{=}t) = \sum_{s_i} p(s_i \mid t) \log \frac{p(t \mid s_i)}{p(t)}$ is the specific information that source $S_i$ provides about the particular outcome $T{=}t$.
The remaining atoms are derived from redundancy: unique information is $U(S_i; T) = I(S_i; T) - I_\cap(S_1; S_2; T)$, and synergy is the residual:
\begin{equation}
  CI(S_1; S_2; T) = I(S_1, S_2; T) - I_\cap(S_1; S_2; T) - U(S_1; T) - U(S_2; T).
\end{equation}

For three or more sources, the decomposition is defined over a lattice of antichains ordered by refinement.
Each node in the lattice corresponds to a distinct mode of information sharing, and the partial information atoms are obtained via M\"{o}bius inversion on this lattice.
In the three-source case used here (SobelX, SobelY, Center $\to$ $\Delta s(t{+}1)$), the lattice yields 18 atoms; we report the full three-way redundancy $I_\cap(S_1; S_2; S_3; T)$, individual unique terms, and the full three-way synergy $CI(S_1; S_2; S_3; T)$.
All variables were discretized into 8 bins via quantile binning.
Computation was performed using the \texttt{PID\_WB} estimator from the \texttt{dit} library.

\section{Supplementary Analyses}
\label{app:analysis}

\subsection{Transfer Entropy Discretization Sensitivity}

All information-theoretic analyses in the main text used 8-bin quantile discretization. Because transfer entropy estimates can depend on the discretization of continuous state variables, we examined whether the qualitative TE asymmetry pattern reported in Figure~\ref{fig:te_fields} was sensitive to the number of quantile bins. Specifically, we repeated the early post-damage TE analysis using 4-, 8-, and 16-bin discretization while keeping all other analysis parameters fixed, including the history length $k=1$, the early post-damage time window, and the construction of the TE asymmetry vector field.

\begin{figure}[!htb]
  \centering
  \includegraphics[width=\linewidth]{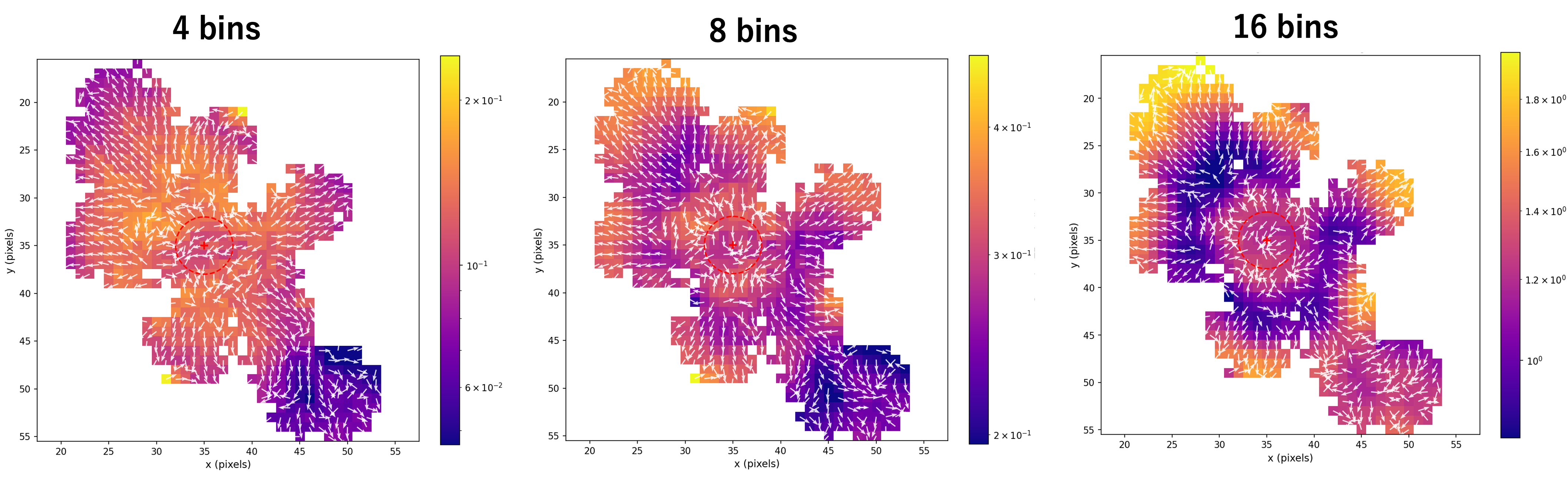}
  \caption{Transfer-entropy asymmetry vector fields computed with different quantile discretizations.
    The same early post-damage response was analyzed using 4-, 8-, and 16-bin quantile discretization.
    Although the absolute TE magnitude changes with the number of bins, the qualitative spatial organization of the vector field is preserved across discretizations, including inward-oriented vectors around the damaged region and outward-oriented propagation in more distal regions.
    The 8-bin discretization used in the main analysis was therefore retained as a compromise between resolution and sample efficiency.\label{fig:te_bins}}
\end{figure}

The resulting vector fields are shown in Figure~\ref{fig:te_bins}. As expected, the absolute TE magnitude varied with the number of bins, reflecting the dependence of empirical TE estimates on discretization resolution and sample efficiency. However, the qualitative spatial organization of the asymmetry field was preserved across all three discretizations. In particular, all conditions showed inward-oriented asymmetry vectors around the damaged region and outward-oriented vectors in more distal regions. Thus, the main conclusion of the TE analysis---that early repair is accompanied by a spatially differentiated inward/outward TE asymmetry pattern---does not depend on the specific choice of 8 bins.

We therefore retained 8-bin quantile discretization in the main analysis as a compromise between resolution and sample efficiency. Fewer bins provide more stable empirical distributions but may collapse relevant state-dependent variation, whereas larger numbers of bins increase resolution but can make the joint distributions used for TE estimation sparse.

\subsection{Additional Target Morphologies}
\begin{figure}[!htb]
  \centering
  \includegraphics[width=\linewidth]{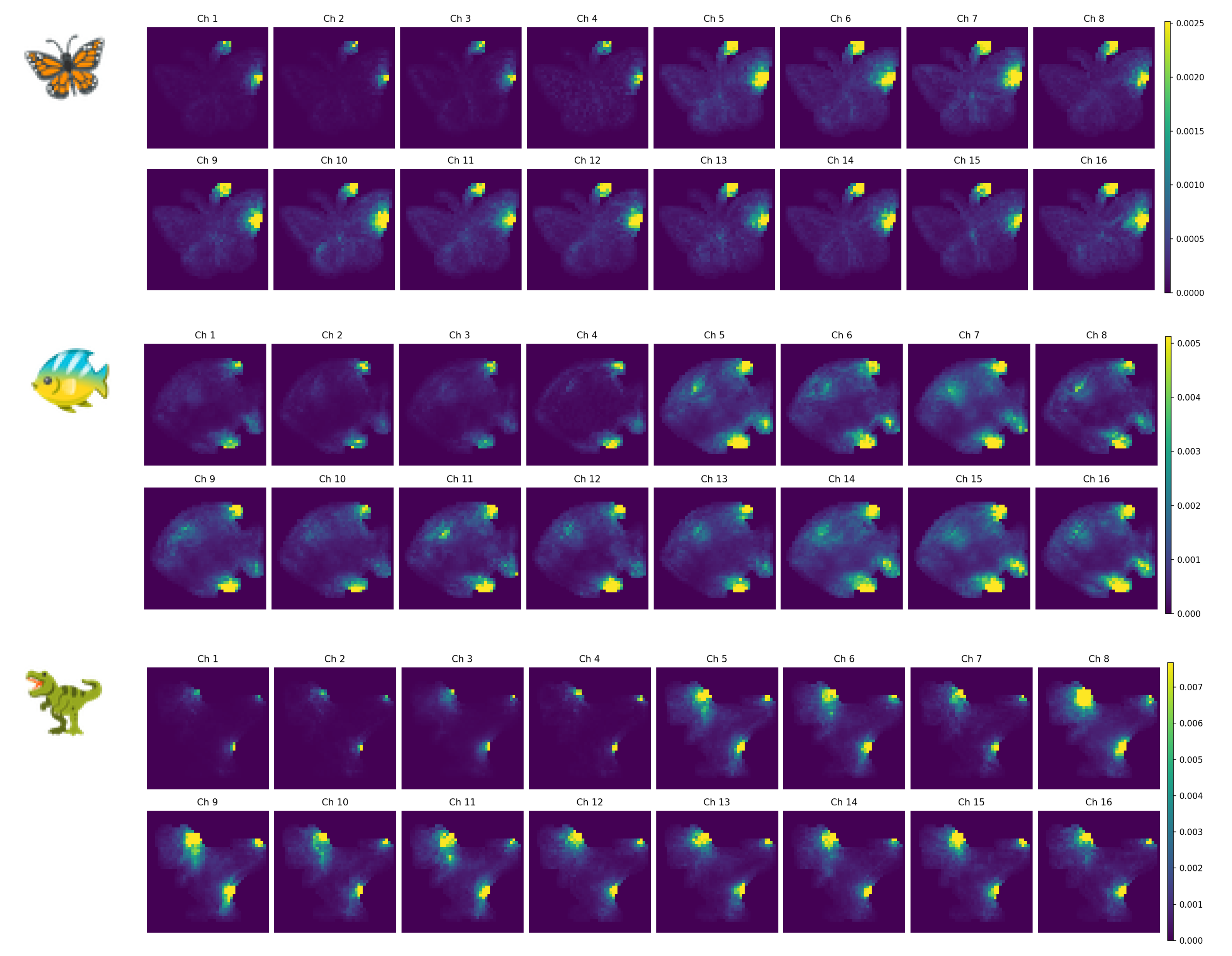}
  \caption{
Temporal activity maps across channels for additional target morphologies.
Rows show GNCA models trained to maintain butterfly, fish, and T-Rex target morphologies.
For each morphology, temporal activity was computed separately for all 16 channels during undamaged steady-state operation.
Across all three target morphologies, activity is not uniformly distributed over the organism but concentrated in spatially localized regions.
These high-activity regions appear at similar spatial locations across channels, indicating channel-invariant fluctuation cores analogous to those observed in the lizard morphology.\label{fig:additional_activity}}
\end{figure}

To examine whether the spatial organization of fluctuations was specific to the lizard morphology, we repeated the steady-state activity analysis for three additional target morphologies: butterfly, fish, and T-Rex. As shown in Figure~\ref{fig:additional_activity}, temporal activity was spatially heterogeneous in all three cases. Rather than being uniformly distributed across the organism, high-activity regions were concentrated in a small number of localized areas. These regions tended to appear at similar spatial locations across channels, suggesting channel-invariant fluctuation-core structures similar to those observed in the lizard model.

\begin{figure}[!htb]
  \centering
  \includegraphics[width=\linewidth]{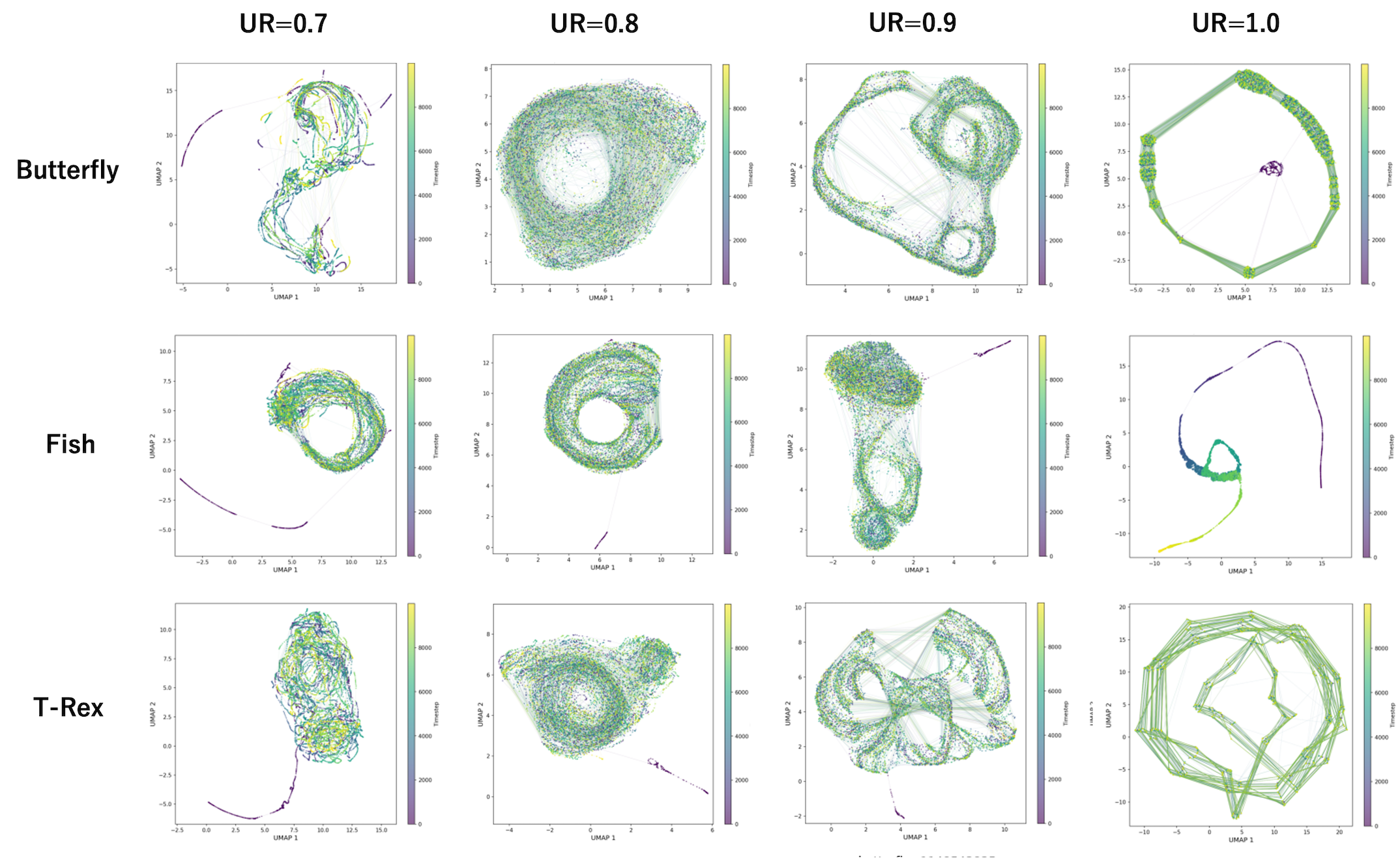}
  \caption{
UMAP embeddings of system-level state trajectories for additional target morphologies.
Rows show butterfly, fish, and T-Rex target morphologies, and columns show update rates $\mathrm{UR}=0.7$, $0.8$, $0.9$, and $1.0$.
Each panel shows a two-dimensional UMAP projection of the high-dimensional trajectory during undamaged steady-state operation, with color indicating time progression.
Across morphologies, higher update rates produced compact and structured trajectories, including loop-like or folded paths at $\mathrm{UR}=0.8$ and $0.9$ and more regular periodic patterns in several $\mathrm{UR}=1.0$ cases.
These results suggest that structured recurrent collective trajectories are observed across multiple target morphologies, although the detailed geometry of the projected trajectories varies with morphology and update rate.\label{fig:additional_umap}
}
\end{figure}

\begin{figure}[!htb]
  \centering
  \includegraphics[width=\linewidth]{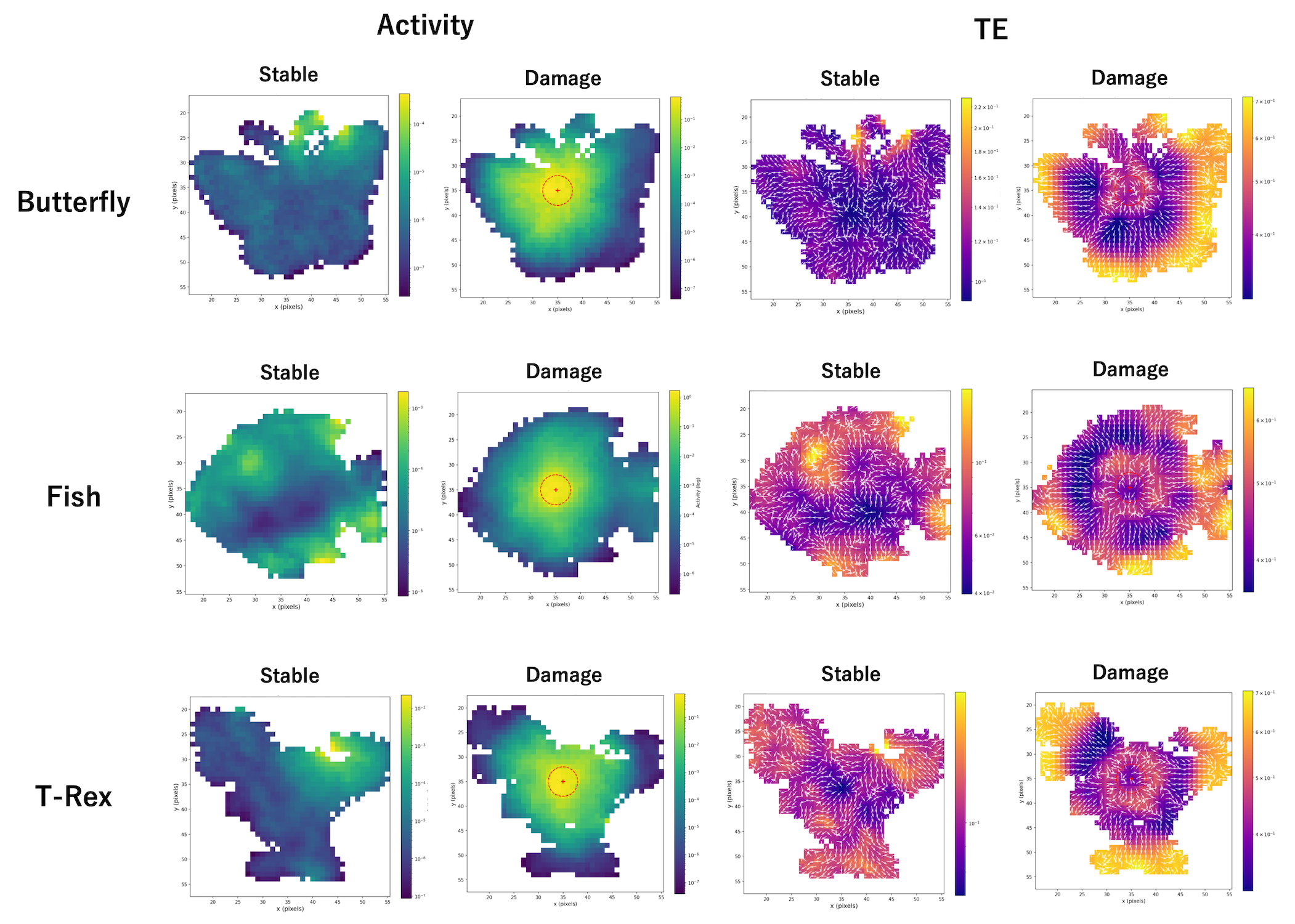}
  \caption{Activity and transfer-entropy asymmetry fields for additional target morphologies.
Rows show butterfly, fish, and T-Rex target morphologies.
For each morphology, the left two columns show mean activity $|\Delta s|$ during undamaged steady-state operation and the early post-damage response, respectively.
The right two columns show the corresponding TE asymmetry vector fields under the same stable and damage conditions.
Background color indicates TE magnitude, and white arrows indicate the normalized TE asymmetry vector field constructed from outgoing minus incoming TE between neighboring cells.
Across morphologies, localized damage induces a broad increase in activity around the damaged region and reorganizes the TE asymmetry field, with inward-oriented vectors near the damage site and outward-oriented vectors in more distal regions.
These results indicate that the spatially differentiated TE response observed in the lizard morphology is qualitatively reproduced in additional target morphologies.\label{fig:additional_te}
}
\end{figure}

We further examined the collective state trajectories of these additional morphologies using UMAP projections (Figure~\ref{fig:additional_umap}). Across morphologies, high update rates produced compact, structured recurrent trajectories, indicating that the emergence of recurrent collective dynamics was not specific to the lizard target.

We next examined whether the spatially differentiated TE response observed in the lizard model also appeared in the additional morphologies. Figure~\ref{fig:additional_te} shows activity maps and TE asymmetry vector fields for the butterfly, fish, and T-Rex targets under stable and early post-damage conditions. In all three morphologies, damage produced a broad increase in activity around the perturbed region and reorganized the TE asymmetry field. Although the detailed geometry varied with morphology, the post-damage fields consistently showed inward-oriented vectors near the damage site together with outward-oriented vectors in more distal regions. Thus, the spatially differentiated inward/outward TE response was qualitatively reproduced across multiple target morphologies.

\begin{figure}[!htb]
  \centering
  \includegraphics[width=\linewidth]{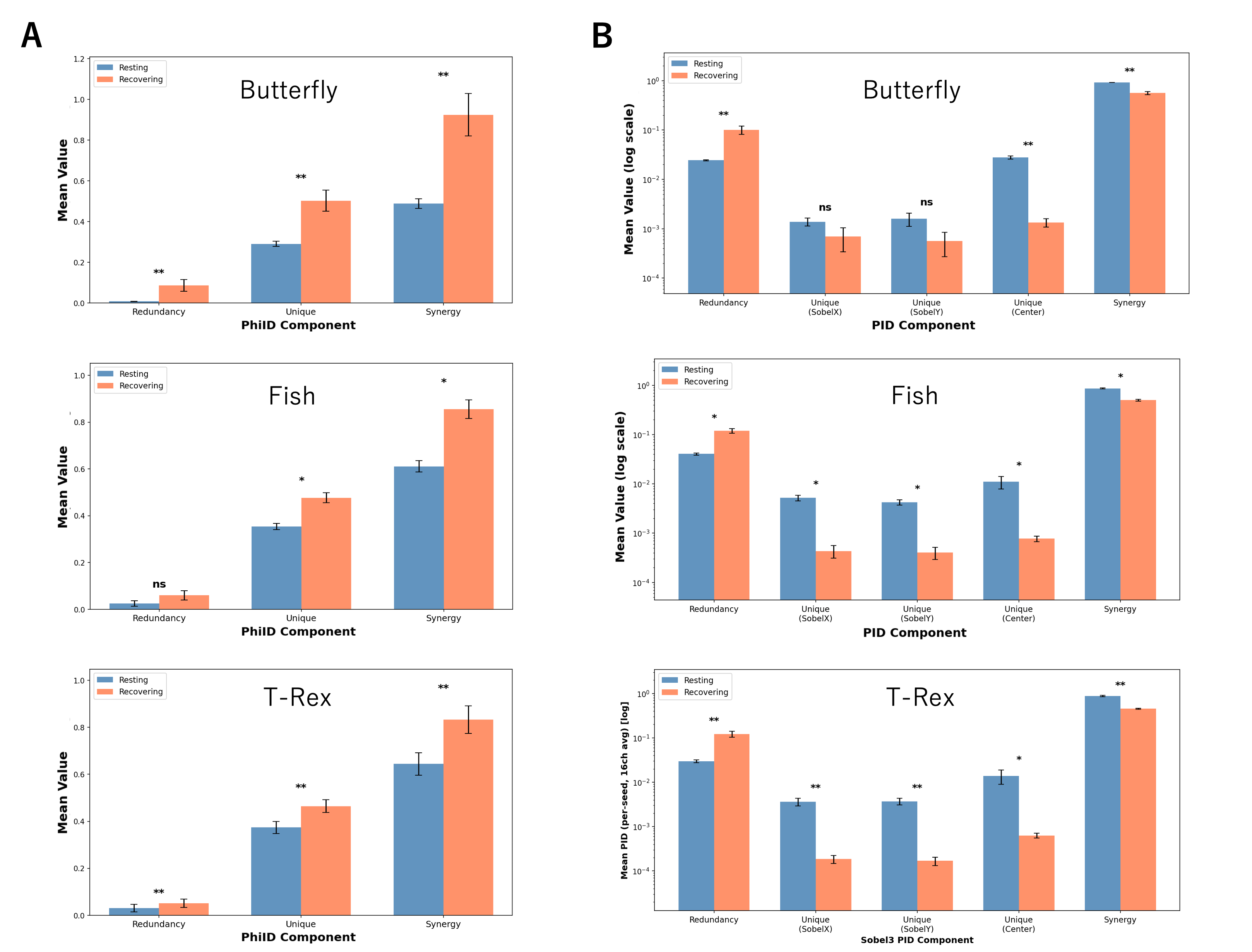}
  \caption{Information-decomposition analyses for additional target morphologies.
(A) Inter-channel $\Phi$ID components under resting and recovering conditions for butterfly, fish, and T-Rex morphologies.
For all morphologies, synergy is the dominant component, followed by unique information, while redundancy remains comparatively small.
Recovery increases the overall level of inter-channel information decomposition, broadly reproducing the pattern observed in the lizard morphology.
(B) Spatial PID components under resting and recovering conditions for the same morphologies.
Across all targets, recovery is associated with increased redundancy and reduced synergy, consistent with a shift toward redundancy-enhanced spatial coordination during repair.
Bars indicate mean values and error bars indicate variability across independently trained seeds.
Asterisks indicate statistical significance of resting--recovering comparisons computed at the seed level; ns indicates non-significant comparisons.\label{fig:additional_pid_phiid}
}
\end{figure}

Finally, we examined whether the information-decomposition signatures observed in the lizard morphology were also present in the additional target morphologies.
Figure~\ref{fig:additional_pid_phiid} summarizes inter-channel $\Phi$ID and spatial PID analyses for the butterfly, fish, and T-Rex targets.
The $\Phi$ID results showed the same overall hierarchy across morphologies, with synergy as the dominant component, followed by unique information, and comparatively small redundancy.
During recovery, all morphologies showed an increase in the overall magnitude of the $\Phi$ID components, suggesting intensified inter-channel coupling during repair.

The spatial PID results also qualitatively reproduced the main pattern observed in the lizard morphology.
In all three additional morphologies, recovery was associated with increased redundancy and decreased synergy relative to the resting condition.
Thus, although the detailed magnitudes varied across target shapes, the information-decomposition analyses support the view that the shift toward redundancy-enhanced spatial coordination during repair is not specific to the lizard morphology.

For these comparisons, values were first averaged within each independently trained model, and statistical comparisons were performed across independently trained models.

\bibliographystyle{unsrtnat}
\bibliography{references}

\end{document}